\documentclass{article}
\usepackage{paper}

\title{
Tuple-oriented Compression for Large-scale \\\
Mini-batch Stochastic Gradient Descent
}
\author{
  {\normalsize
    Fengan Li{$^\dag$} \thanks{These authors are currently at Google} \hspace{1mm}
    \ \ Lingjiao Chen{$^\dag$} \hspace{1mm}
    \ \ Yijing Zeng{$^\dag$} \hspace{1mm}
    \ \ Arun Kumar {$^\S$} \hspace{1mm}
  } \\
  { \normalsize
    \ \ Jeffrey F. Naughton{$^\dag$} \footnotemark[1] \hspace{1mm}
    \ \ Jignesh M. Patel{$^\dag$} \hspace{1mm}
	\ \ Xi Wu {$^\dag$} \footnotemark[1] \hspace{1mm}
  } \\
  {\normalsize
    {$^\dag$} University of Wisconsin-Madison \hspace{1mm}
    {$^\S$} University of California, San Diego
  } \\
  \small{
    {$^\dag$}\{fengan, lchen, yijingzeng, naughton, jignesh, xiwu\}@cs.wisc.edu \hspace{1mm}
    {$^\S$} arunkk@eng.ucsd.edu	
  }
}

\begin{document}
\maketitle



\begin{abstract}
Data compression is a popular technique for improving the efficiency of data processing workloads such as SQL queries and more recently, machine learning (ML) with classical batch gradient methods. But the efficacy of such ideas for mini-batch stochastic gradient descent (MGD), arguably the workhorse algorithm of modern ML, is an open question. MGD's unique data access pattern renders prior art, including those designed for batch gradient methods, less effective. We fill this crucial research gap by proposing a new lossless compression scheme we call tuple-oriented compression (TOC) that is inspired by an unlikely source, the string/text compression scheme Lempel-Ziv-Welch, but tailored to MGD in a way that preserves tuple boundaries within mini-batches. We then present a suite of novel compressed matrix operation execution techniques tailored to the TOC compression scheme that operate directly over the compressed data representation and avoid decompression overheads. An extensive empirical evaluation with real-world datasets shows that TOC consistently achieves substantial compression ratios by up to 51x and reduces runtimes for MGD workloads by up to 10.2x in popular ML systems.
\end{abstract}


\section{Introduction}
\label{sec:intro}
\begin{figure*}[th!]
\centering
    \includegraphics[width=0.97\linewidth]{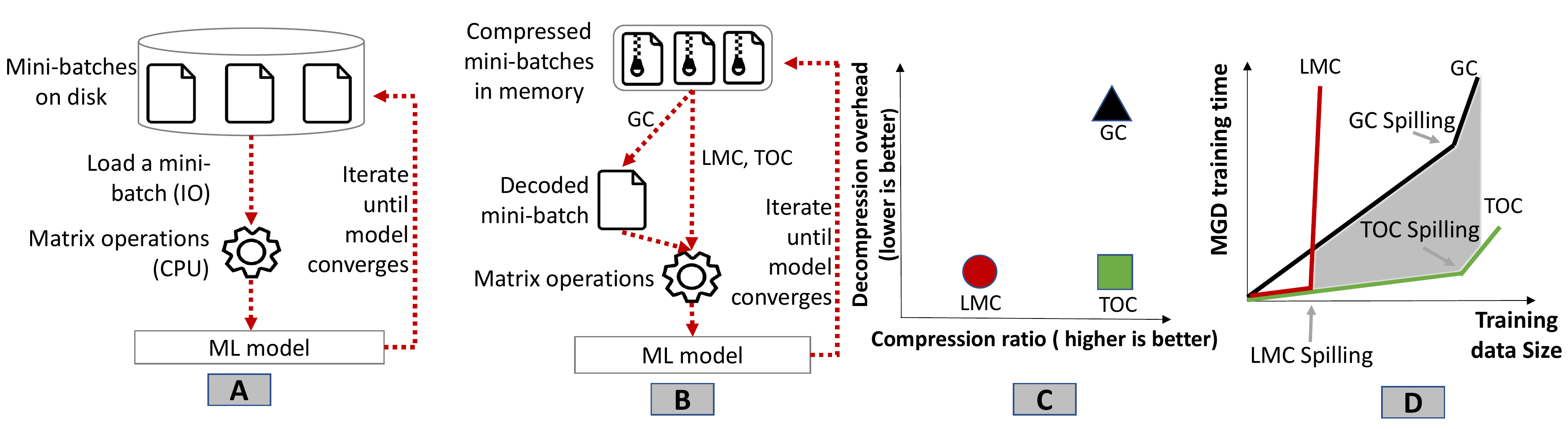}
\vspace{2mm}
    \caption{\textbf{A:} If no compression is used to train ML models on large datasets that cannot fit into memory, loading a mini-batch (IO time) from disk
    can be significantly more expensive than matrix operations (CPU time) performed on the mini-batch for MGD. \textbf{B:} One typically uses a compression
    scheme to compress mini-batches so that they can fit into memory. For general compression schemes (GC), a mini-batch has to be decoded before any computation can be carried out. For light-weight matrix compression schemes (LMC) and our proposed tuple-oriented compression (TOC), matrix
    operations can directly operate on the encoded output without decompression overheads. \textbf{C:} TOC has compression ratios comparable to GC. Similar 
    to LMC, matrix operations can directly operate on the TOC output without decoding the mini-batch. \textbf{D:} Since TOC has good compression ratios and
    no decompression overheads, it reduces the MGD training time especially on large datasets. For small datasets, TOC has comparable performance
    to LMC. Note that MGD training time grows sharply once the data is spilled to disk.}
\label{fig:intro_overview}
\end{figure*}

Data compression is a popular technique for improving the efficiency of data processing workloads such as SQL queries over compressed 
databases~\cite{abadi2006integrating,li2013bitweaving,wesley2014leveraging,elgohary2016compressed,wang2017experimental} and more recently, machine learning with classical batch gradient methods~\cite{elgohary2016compressed}. However, to the best of our knowledge, there is no such study of data compression for \textbf{mini-batch stochastic gradient descent (MGD)}
~\cite{hogwild,wu2017bolt,bismarck,kaoudi2017cost,qin2017scalable}, which is known for its fast convergence rate and statistical stability, and is arguably
the workhorse algorithm~\cite{ruder2016overview,hinton2012neural} of modern ML. This research gap is getting more crucial as training dataset sizes in ML keep growing~\cite{chelba2013one,russakovsky2015imagenet}. For example, if no compression is used to train ML models on large datasets that cannot fit into memory capacity or even distributed memory capacity, disk IO time becomes a significant overhead~\cite{elgohary2016compressed,yu2012large} for MGD. Figure~\ref{fig:intro_overview}A highlights this issue in more detail.

Despite the need for a good data compression scheme to improve the efficiency of MGD workloads, unfortunately, the main existing data compression schemes
designed for general data files or batch gradient methods are not a good fit for the data access pattern of MGD.
Figure~\ref{fig:intro_overview}B highlights these existing solutions. For examples,
\textbf{general compression schemes (GC)} such as Gzip and Snappy are designed for general data files. GC typically has good
compression ratios on mini-batches; however, a mini-batch has to be decompressed before any computation can be carried out, and the decompression
overhead is significant~\cite{elgohary2016compressed} for MGD. \textbf{Light-weight matrix compression schemes (LMC)} include
classical methods such as compressed sparse row~\cite{saad2003iterative} and value indexing~\cite{kourtis2008optimizing} and more recently, a state-of-the-art technique called compressed linear algebra~\cite{elgohary2016compressed}. LMC is suitable for batch gradient methods because the compression ratio of LMC is
satisfactory on the whole dataset and matrix operations can directly operate on the encoded output without decompression
overheads. Nevertheless, the compression ratio of LMC on mini-batches is not as good as GC in general, which makes it less attractive for MGD.

In this paper, we fill this crucial research gap by proposing a lossless matrix compression scheme called \textbf{tuple-oriented compression (TOC)},
whose name is based on the fact that tuple boundaries (i.e., boundaries between columns/rows in the underlying tabular data) are 
preserved. Figure~\ref{fig:intro_overview}C highlights the advantage of TOC over existing compression schemes. TOC has both good compression ratios
on mini-batches and no decompression overheads for matrix operations, which are the main operations executed by MGD on compressed data.
Orthogonal to existing works like GC and LMC, TOC takes inspirations from an 
unlikely source---a popular string/text compression scheme Lempel-Ziv-Welch (LZW)~\cite{welch1984technique,ziv1977universal,ziv1978compression}---and builds a compression scheme with compression ratios comparable to Gzip on mini-batches.
In addition, this paper also proposes a suite of compressed matrix operation execution techniques, which are tailored to the TOC compression scheme, that {\em operate directly over the compressed data representation} and avoid decompression overheads. Even for a small dataset that fits into memory, these
compressed execution techniques are often faster than uncompressed execution techniques because they can reduce computational redundancies in matrix operations. Collectively, these techniques present a fresh perspective that weaving together ideas from databases, text processing, and ML can achieve substantial efficiency gains for popular MGD-based ML workloads. Figure~\ref{fig:intro_overview}D highlights the effect of TOC in reducing the MGD runtimes, especially on large datasets.

TOC consists of three components at different layers of abstraction: sparse encoding, logical encoding, and physical encoding. All these components respect the boundaries between rows and columns in the underlying tabular data so that matrix operations
can be carried out on the encoded output directly. Sparse encoding uses the well-known sparse row technique~\cite{saad2003iterative} as a starting point. Logical encoding uses a prefix tree encoding algorithm, which is based on the LZW compression scheme, to further compress matrices. Specifically, we notice that there are sequences of column values which are repeating across matrix rows. Thus, these repeated sequences can be stored in a prefix tree and each tree node represents a sequence. The occurrences
of these sequences in the matrix can be encoded as indexes to tree nodes to reduce space. Note that we only need
to store the encoded matrix and the first layer of the prefix tree as encoded outputs, as the prefix tree can be rebuilt
from the encoded outputs if needed. Lastly, physical encoding encodes integers and float numbers efficiently.

We design a suite of compressed execution techniques that operate directly over the compressed data representation
without decompression overheads for three classes of matrix operations. These matrix operations are used by MGD to train popular ML models such as Linear/Logistic regression, Support vector machine, and Neural network. These compressed execution techniques only need to scan the 
encoded table and the prefix tree at most once. Thus, they are fast, especially when TOC exploits significant redundancies. For example, right multiplication (e.g.,
matrix times vector) and left multiplication (e.g., vector times matrix) can be computed with one scan
of the encoded table and the prefix tree. Lastly, since these compressed execution
techniques for matrix operations differ drastically 
from the uncompressed execution techniques, we provide mathematical analysis to prove the correctness of these compressed techniques.

To summarize, the \textbf{main contributions} of this paper are:
\begin{enumerate}
\item To the best of our knowledge, this is the first work to study lossless compression techniques to reduce the memory/storage footprints and runtimes for mini-batch stochastic gradient descent (MGD), which is the workhorse algorithm of modern ML.
We propose a lossless matrix compression scheme called tuple-oriented compression (TOC) with compression ratios comparable to Gzip on mini-batches.
\item We design a suite of novel compressed matrix operation execution techniques tailored to the TOC compression scheme that directly operate over the compressed data representation and avoid decompression overheads for MGD workloads.
\item We provide a formal mathematical analysis to prove the correctness of the above compressed
matrix operation execution techniques.
\item We perform an extensive evaluation of TOC compared to seven compression schemes on six real
datasets. Our results show that TOC consistently achieves substantial compression ratios by up to 51x. Moreover, TOC reduces MGD runtimes for three popular ML models by up to 5x compared to the state-of-the-art compression schemes and by up to 10.2x compared to the encoding methods in some popular ML systems (e.g., ScikitLearn~\cite{scikit-learn}, Bismarck~\cite{bismarck} and TensorFlow~\cite{abadi2016tensorflow}). An integration of TOC into Bismarck also confirmed that TOC can greatly benefit MGD performance in ML systems.
\end{enumerate}

The remainder of this paper proceeds as follows: \S~\ref{sec:preliminaries} presents some required background information. \S~\ref{sec:encoding} explains our TOC compression scheme, while
\S~\ref{sec:la} presents the techniques to execute matrix operations on the compressed data. \S~\ref{sec:expts} presents
the experimental results and \S~\ref{sec:discuss} discusses TOC extensions. \S~\ref{sec:related} presents related work, and we conclude in \S~\ref{sec:conclusion}.

\section{Background}\label{sec:preliminaries}
In this section, we discuss two important concepts: ML training in the
generalized setting and data compression.

\subsection{Machine Learning Training}
\subsubsection{Empirical Risk Minimization}
We begin with a description of ML training in the generalized setting based on standard ML texts~\cite{SSSSS10,shai}. Formally, we have a hypothesis space $\mathcal{H}$,
an instance set $\mathcal{Z}$, and a loss function $\ell: \mathcal{H} \times \mathcal{Z} \mapsto \bbR$.
Given a training set
$\mathcal{S} = \{z_1, z_2, ..., z_n\}$ which are $n$ i.i.d. draws based on a distribution $\mathcal{D}$ on $\mathcal{Z}$, and a hypothesis
$h \in \mathcal{H}$, our goal is to minimize
the {\em empirical risk} over the training set $\mathcal{S}$ defined as
\begin{equation} \label{eq:general}
 L_{\mathcal{S}}(h) = \frac{1}{n}\sum_{i=1}^n \ell(h, z_i).
\end{equation}

Many ML models including Logistic/Linear regression, Support vector machine, and Neural network fit into this generalized setting~\cite{SSSSS10}.

\subsubsection{Gradient Descent}
ML training can be viewed as the process to find the optimal $\hat{h} \in \mathcal{H}$ such that $\hat{h} = \textrm{argmin}\  L_{\mathcal{S}}(h).$ This is essentially an optimization problem, and gradient descent is a common and established class of algorithms for solving this problem. There
are three main variants of gradient descent: batch gradient descent, stochastic gradient descent,
and mini-batch stochastic gradient descent.

\noindent \textbf{Batch Gradient Descent (BGD)}. BGD uses all the training data to compute the gradient and update $h$ per iteration.

\noindent \textbf{Stochastic Gradient Descent (SGD)}. SGD uses a single tuple to compute the gradient and update $h$ per iteration.

\noindent \textbf{Mini-batch Stochastic Gradient Descent (MGD)}.  MGD uses a small batch of tuples (typically tens or  hundreds of tuples) to compute the gradient and update $h$ per iteration:

\begin{equation} \label{equation:gradient_descent}
h^{(t)} \leftarrow h^{(t-1)} - \lambda \frac{1}{|B^t|}\sum_{z \in B^t}\frac{\partial \ell(h, z)}{\partial h},
\end{equation}
where $B^t$ is the current $t$-th mini-batch we visit, $z$ is a tuple from $B^t$, and $\lambda$ is the learning rate.

Note that MGD can cover the spectrum of
gradient descent methods by setting the mini-batch size $|B^t|$. For examples, MGD morphs into SGD and BGD by setting $|B^t| = 1$ and $|B^t| = |\mathcal{S}|$, respectively.

\begin{figure}[th!]
\centering
    \includegraphics[width=0.97\linewidth]{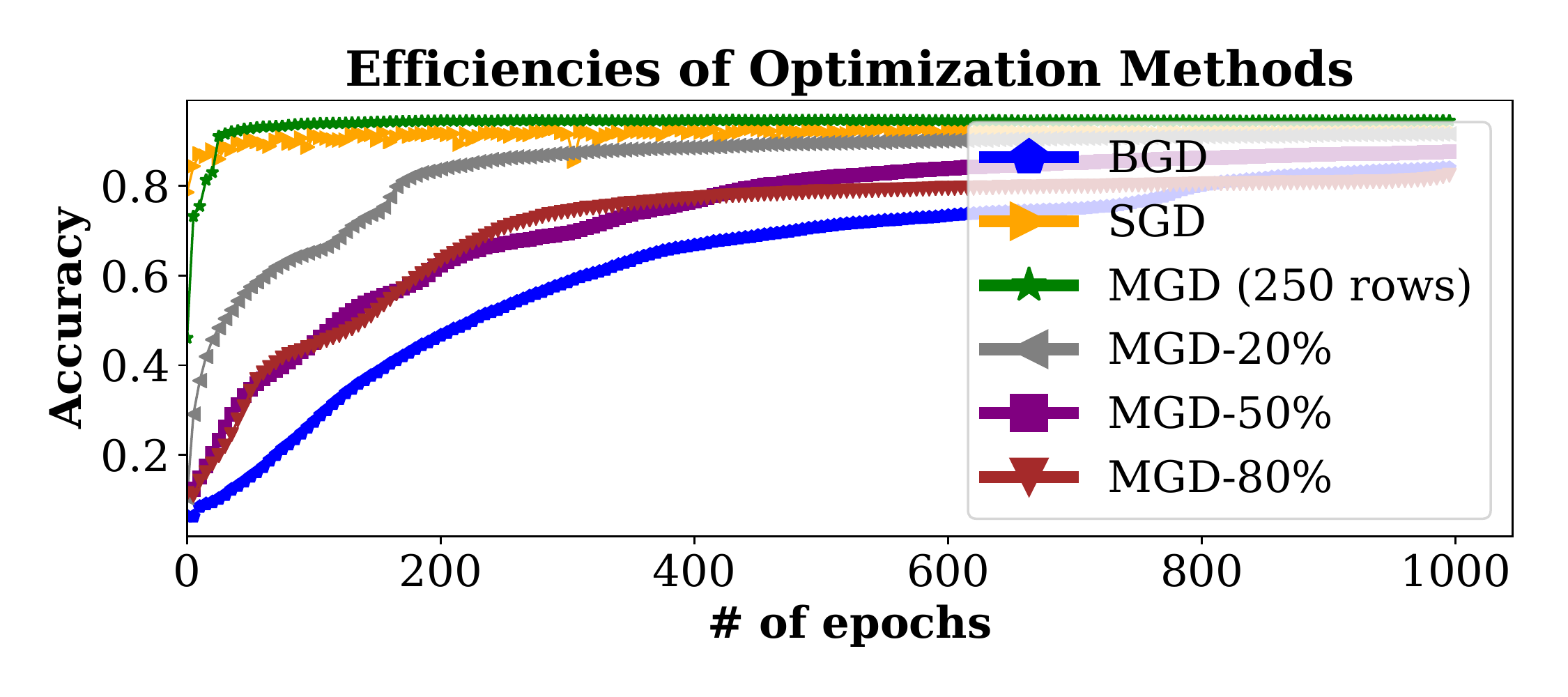}
\vspace{2mm}
    \caption{Optimization efficiencies of BGD, SGD, and MGD for training a neural network with one hidden layer (no convolutional layers) on Mnist. MGD (250 rows)
    has 250 rows in a mini-batch. MGD-20\%, MGD-50\%, and MGD-80\% has 20, 50, and 80 percent of rows of the
    whole dataset in each mini-batch, respectively.}
\label{fig:optimization_methods_efficiency}
\vspace{-3mm}
\end{figure}

MGD gains its popularity due to its fast convergence rate and statistical stability. It typically requires fewer epochs (the whole pass over a dataset is an epoch) to
converge than BGD and is more stable than SGD~\cite{ruder2016overview}. Figure~\ref{fig:optimization_methods_efficiency} illustrates the optimization efficiencies of these gradient descent variants, among which MGD with hundreds of rows in a mini-batch achieves the best balance between fast convergence rate and statistical stability. Thus, in this paper, we focus on MGD with mini-batch size ranging from tens to hundreds of tuples.

\subsubsection{Shuffle Once v.s. Shuffle Always}
The random sampling for tuples in SGD/MGD is typically done without replacement, which is achieved by shuffling the dataset before an epoch~\cite{bengio2012practical}. However, shuffling data at every epoch (shuffle always) is expensive and incurs a high overhead. Thus, we follow the standard technique of shuffling once~\cite{bismarck,wu2017bolt,bengio2012practical} (i.e., shuffling the data once upfront) to improve the ML training efficiency.

\subsubsection{Core Matrix Operations for Gradient Descent}
The core operations, which dominate the CPU time, for using gradient descent to optimize many ML models (e.g., Linear/Logistic regression, Support vector machine, and Neural network) are matrix
operations~\cite{elgohary2016compressed}. We illustrate this point using an example of Linear regression, and summarize the core matrix operations for these ML models in Table~\ref{tab:matrix_operations_for_gradient_descent}.
\begin{table}[th!]
 \vspace{1mm}
 \caption{The core matrix operations when using gradient descent to optimize popular ML models. $A = [x_1^T; x_2^T; ...; x_{|B|}^T]$ is
         a batch of data for updating models where $(x_i, y_i) \in B$. $v$ and $M$ are either ML model parameters or intermediate results for
         computing gradients. We use logistic loss for Logistic regression, hinge loss for Support vector machine, and mean squared loss for Linear regression and Neural network. For the sake
         of simplicity, our neural network structure has a feed forward structure with a single hidden layer.}
 \vspace{-1mm}
 \label{tab:matrix_operations_for_gradient_descent}
 \centering
       \begin{tabular}{|c||c|c|c|c|}
         \hline
         \textbf{ML models} & $A \cdot v$ & $v \cdot A$ & $A \cdot M$ & $M \cdot A$ \\
         \hline
         \hline
         Linear regression & $\checkmark$ & $\checkmark$ & & \\
         \hline
         Logistic regression      & $\checkmark$ &  $\checkmark$ &        &                 \\
         \hline
         Support vector machine      & $\checkmark$ &  $\checkmark$ &         &                 \\
         \hline
         Neural network     & & & $\checkmark$ & $\checkmark$                          \\
         \hline
\end{tabular}

 \end{table}
\vspace{1mm}

\noindent \textbf{Example}. Consider a supervised classification algorithm Linear regression
where $\mathcal{Z} = \mathcal{X} \times \mathcal{Y}$, $\mathcal{X} \subseteq \bbR^d$, $\mathcal{Y} = \bbR$,
$\mathcal{H} = \bbR^d$, and $\ell(h, z) = \frac{1}{2} (y-x^Th)^2$. Let matrix $A = [x_1^T; x_2^T; ...; x_{|B|}^T]$, vector $Y = [y_1; y_2; ...; y_{|B|}]$, then the aggregated 
gradients of the loss function is:
\begin{equation} \label{equation:loss_function}
\sum_{ z \in B} \frac{\partial \ell(h, z)}{\partial h} = \sum_{(x, y) \in B} (x^Th - y)x
  = ((Ah-Y)^TA)^T.
\end{equation}
Thus, there are two core matrix operations---matrix times vector and vector times matrix---to compute
Equation~\ref{equation:loss_function}.

\subsection{Data Compression}
Data compression, also known as source coding, is an important technique to reduce data sizes.
There are two main components in a data compression scheme, an encoding process that encodes the
data into coded symbols (hopefully with fewer bits), and a decoding process that reconstructs the
original data from the compressed representation.

Based on whether the reconstructed data differs from the original data, data compression schemes
usually can be classified into {\em lossless} compression or {\em lossy} compression. In this paper,
we propose a lossless compression scheme called tuple-oriented compression (TOC) which is inspired by a classical lossless string/text compression scheme that has both gained academic influence and industrial popularity,
Lempel-Ziv-Welch~\cite{welch1984technique,ziv1977universal,ziv1978compression}. For examples, Unix file compression utility\footnote{ncompress.sourceforge.net} and GIF~\cite{wiggins2001image} image format are based on LZW.

\section{Tuple-oriented Compression}\label{sec:encoding}
\begin{figure*}[th!]
\centering
    \includegraphics[width=0.97\linewidth]{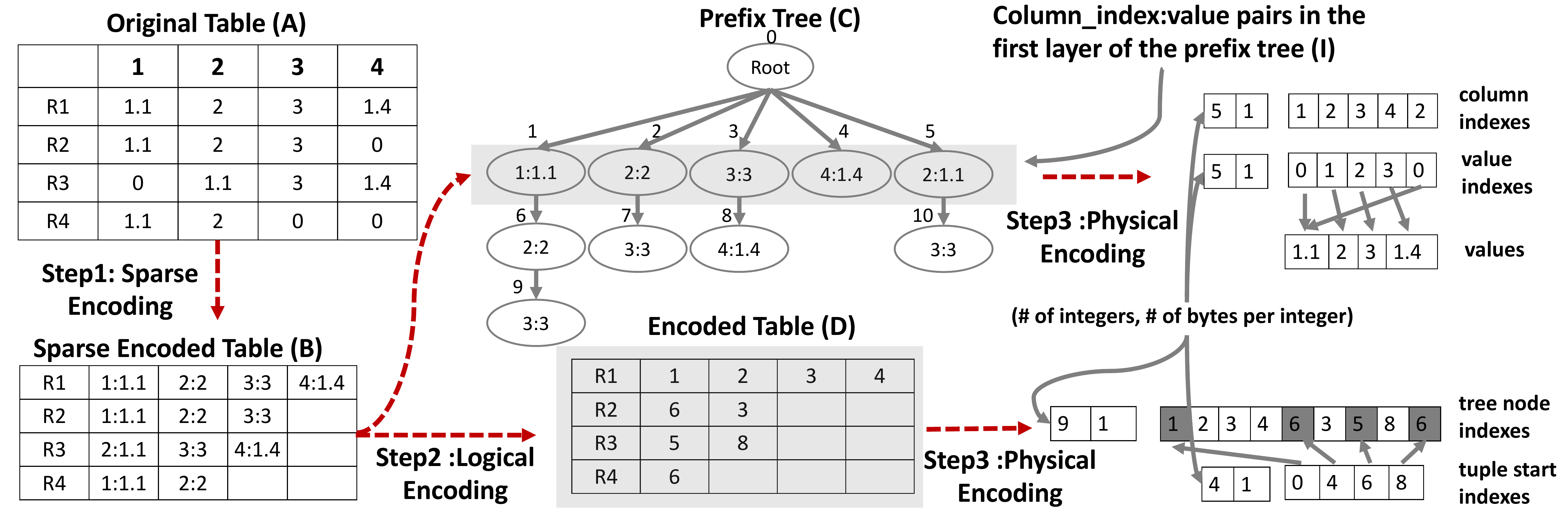}
\vspace{2mm}
    \caption{A running example of the TOC encoding process. TOC has three components: sparse encoding,
    logical encoding, and physical encoding. The red dotted lines connect these components.
    Sparse encoding encodes the original table \textbf{A} to the sparse encoded table \textbf{B}.
    Logical encoding encodes \textbf{B} to the encoded table \textbf{D}. It also outputs \textbf{I}, which
    is the column\_index:value pairs in the first layer of the prefix tree \textbf{C}. Physical encoding
    encodes \textbf{I} and \textbf{D} to physical bytes efficiently.}
\label{fig:encoding-overview}
\vspace{-3mm}
\end{figure*}

In this section, we introduce our tuple-oriented compression (TOC). The goal of TOC is to (1) compress a mini-batch as much as
possible and (2) preserve the row/column boundaries in the underlying tabular data so that matrix operations can directly operate
on the compressed representation without decompression overheads. Following the popular sparse row technique~\cite{saad2003iterative}, we
use sparse encoding as a starting point, and introduce two new techniques: logical encoding and physical encoding.
Figure~\ref{fig:encoding-overview} demonstrates a running example of the encoding process.

For sparse encoding, we ignore the zero values and then prefix each non-zero value with its column index. We call the 
value with its column index prefix as \emph{column\_index:value pair}. For example, in Figure~\ref{fig:encoding-overview},
tuple R2 - [1.1, 2, 3, 0] is encoded as [1:1.1, 2:2, 3:3], where 1:1.1 is a column\_index:value pair. As a result of sparse encoding, the original 
table (\textbf{A}) in Figure~\ref{fig:encoding-overview} is converted to the sparse encoded table (\textbf{B}).

\subsection{Logical Encoding} \label{sec:logical_encoding}
The sparse encoded table (e.g., \textbf{B} in Figure~\ref{fig:encoding-overview}) can be further compressed
logically.
The key idea is that there are repeating sequences of column\_index:value pairs across tuples in the table. For example,
R2 and R4 in the table \textbf{B} both have the same sequence [1:1.1, 2:2]. Thus, these occurrences of
the same sequence can be encoded as the same index pointing to a dictionary entry, which represents the original sequence.
Since many of these sequences have common prefixes, a prefix tree is used to store all the sequences. Finally, each original tuple is encoded as a vector of indexes pointing to prefix tree nodes.

We present the prefix tree structure and its APIs in \S~\ref{sec:api}. In \S~\ref{sec:toc_alg}, we present the actual prefix tree encoding algorithm, including how to dynamically 
build the tree and encode tuples. The comparison between
our prefix tree encoding algorithm and LZW is presented in \S~\ref{sec:lzw_comparisons}.
\subsubsection{Prefix Tree Structure and APIs}
\label{sec:api}
Each node of the prefix tree has an index. Except for the root node, each node stores a column\_index:value pair as its key. Each node also represents a sequence of column\_index:value pairs,
which are obtained by concatenating the keys from the prefix tree root to the node itself. For example, in the prefix tree \textbf{C} in Figure~\ref{fig:encoding-overview}, the left bottom tree node has index 9, 
stores key 3:3, and represents the sequence of column\_index:value pairs [1:1.1, 2:2, 3:3].

The prefix tree supports two main APIs:  \textbf{AddNode} and \textbf{GetIndex}.

\begin{packeditems}
\item $n' = \textbf{AddNode}(n, k)$. This API creates a new prefix tree node which has key $k$ and is 
a child of the tree node with index $n$. It also returns the index of the newly created tree node in $n'$, which is
assigned from a sequence number starting from 0.

\item $n' = \textbf{GetIndex}(n, k)$. This API looks up the tree node which has key $k$ and is a child of the tree node with index $n$.
It also returns the index of the found tree node in $n'$. If there is no such node, it returns -1.
\end{packeditems}

The implementation of \textbf{AddNode} is straightforward. The implementation of \textbf{GetIndex} is more involved, and we use a standard technique reported in~\cite{blelloch2001introduction}. In essence, for each tree node, we create a hash map mapping from its child node 
keys to its child node indexes.

\subsubsection{Prefix Tree Encoding Algorithm}
\label{sec:toc_alg}
Our prefix tree encoding algorithm encodes the sparse encoded table (e.g., \textbf{B} in Figure~\ref{fig:encoding-overview})
to an encoded table (e.g., \textbf{D} in Figure~\ref{fig:encoding-overview}). During the encoding
process, we build a prefix tree (e.g., \textbf{C} in Figure~\ref{fig:encoding-overview}) and each original tuple is encoded as a vector of indexes pointing to prefix tree nodes.
Algorithm~\ref{alg:prefix_tree_encoding} presents the pseudo-code of the algorithm. Figure~\ref{fig:encoding-overview} presents a running example of executing the algorithm and encoding table \textbf{B} to table \textbf{D}.

The prefix tree encoding algorithm has two main phases. In phase \rom{1} (line 5 to line 8 of 
Algorithm~\ref{alg:prefix_tree_encoding}), we initialize the prefix tree
with all the unique ~column\_index:value pairs in the sparse encoded table as the children of the root 
node. 

In phase \rom{2} (line 9 to line 17 of Algorithm~\ref{alg:prefix_tree_encoding}), we leverage the repeated sequences of the tuples so that the same sequence, for example
R2 and R4 in Figure~\ref{fig:encoding-overview} both have the sequence [1:1, 2:2], is encoded as the same index to the prefix tree node.
At its heart, we scan all the tuples to detect if part of the tuple can match a sequence that already exists in the prefix tree and build up the prefix tree along the way. We use the function \textsc{LongestMatchFromTree} in Algorithm~\ref{alg:prefix_tree_encoding}, to find the longest sequence in the prefix tree that matches
the sequence in the tuple $\textbf{t}$ starting from the position $i$.  The function returns the tree node index of the longest match in $n$, and the next matching starting position in $j$. If $j \neq len(\textbf{t})$, we add a new
node to the prefix tree which is the child of the tree node with index $n$ and has key $\textbf{t}[j]$ to capture this new sequence in the tuple $\textbf{t}$. In this way, later tuples can leverage this new sequence. Note that the longest match found is at least of length one because 
we store all the unique column\_index:value pairs as the children of the root node in phase \rom{1}. Table~\ref{tab:encoding_example} gives a running example of executing Algorithm~\ref{alg:prefix_tree_encoding} on table \textbf{B} in Figure~\ref{fig:encoding-overview}.

Our prefix tree encoding and LZW are both linear algorithms in the sense that each
input unit is read at most twice and the operation on each input unit is constant. So the time complexity of Algorithm~\ref{alg:prefix_tree_encoding} is $\mathcal{O}(|\textbf{B}|)$, where $|\textbf{B}|$ is the number of column\_index:value pairs in the sparse encoded table \textbf{B}.

\begin{algorithm}
\caption{Prefix Tree Encoding Algorithm} \label{alg:prefix_tree_encoding}
\begin{algorithmic}[1]
\Function{PrefixTreeEncode}{\textbf{B}}
    \State \textbf{inputs:} sparse encoded table \textbf{B}
    \State \textbf{outputs:} column\_index:value pairs in the first layer
    \Statex of the prefix tree \textbf{I} and encoded table \textbf{D}
    \State Initialize \textbf{C} with a root node with index 0.
    \For {each tuple \textbf{t} in \textbf{B}} \Comment{phase \rom{1}: initialization}
    	\For {each column\_index:value pair \textbf{t}[$i$] in \textbf{t}}
        	\If{\textbf{C.GetIndex}($0, \textbf{t}[i]$) = -1}
            	\State \textbf{C.AddNode}($0, \textbf{t}[i]$)
            \EndIf
        \EndFor
    \EndFor
    
    \For{each tuple \textbf{t} in \textbf{B}} \Comment{phase \rom{2}: encoding}
    	\State $i \gets 0$ \Comment{set the matching starting position}
        \State $\textbf{D[t]} \gets []$ \Comment{initialize as an empty vector}
        \While {$i <$ len(\textbf{t})}
        	\State $(n, j) \gets $ \Call{LongestMatchFromTree}{\textbf{t}, $i$, \textbf{C}}
            \State \textbf{D[t]}.append($n$)
            \If {$j <$ len(\textbf{t})}
            	\State \textbf{C.AddNode}($n$, \textbf{t}[$j$])
            \EndIf
            \State $i \gets j$ 
        \EndWhile
    \EndFor
    \State \textbf{I} $\gets$ first\_layer(\textbf{C})
    \State \textbf{return(I, D)}
\EndFunction
\State
\Function{LongestMatchFromTree}{\textbf{t}, $i$, \textbf{C}}
	\State \textbf{inputs:} input tuple $\textbf{t}$, matching starting position $i$
    \Statex in \textbf{t}, and prefix tree \textbf{C}
    \State \textbf{outputs:} index of the tree node of the longest match
    \Statex $n$ and next matching starting position $j$
    \State $j \gets i$
    \State $n' \gets $ \textbf{C.GetIndex}($0$, \textbf{t}[$j$]) \Comment{matching 1st element}
    \Do
    	\State $n \gets n'$
        \State $j \gets j + 1$ \Comment{try matching the next element}
    	\If {$j <$ len\textbf{(t)}}
        	\State $n' \gets $ \textbf{C.GetIndex}($n$, \textbf{t}[$j$]) \Comment {return -1 if such a tree node does not exist}
        \Else
        	\State $n' \gets -1$ \Comment{reaching the end of tuple \textbf{t}}
        \EndIf
    \doWhile {$n' \neq -1$}
	\State \textbf{return($n$, $j$)}
\EndFunction
  
\end{algorithmic}
\end{algorithm}

\begin{table}[bth]
 \vspace{1mm}
 \caption{
    We show the steps of running Algorithm~\ref{alg:prefix_tree_encoding}
    on table \textbf{B} in Figure~\ref{fig:encoding-overview}.
    We omit the phase \rom{1} of the algorithm, which initializes the prefix tree with nodes:
    1 $\rightarrow$ [1:1.1], 2 $\rightarrow$ [2:2], 3 $\rightarrow$ [3:3], 4 $\rightarrow$ [4:1.4], and 5 $\rightarrow$ [2:1.1],
    where the left side of the arrow is the tree node index and the right side of the arrow is the sequence of column\_index:value pairs represented by the tree node.
    Each entry here illustrates an iteration of the while loop
    (line 12 - line 17) of Algorithm~\ref{alg:prefix_tree_encoding}.
    Column $i$ is the starting position of the tuple that
    we try to match the sequence in the prefix tree. Column \textbf{LMFromTree} shows the index and the
    corresponding sequence of the found longest match by the function \textsc{LongestMatchFromTree}.
    Column \textbf{App} is the appended tree node index for encoding the tuples in table \textbf{B}.
    Column \textbf{AddNode} shows the index and the corresponding sequence of the newly added tree node.}
\label{tab:encoding_example}

 \centering
       \begin{tabular}{|c||c|c|c|c|}
         \hline
           & $i$ & \textbf{LMFromTree} & \textbf{App} & \textbf{AddNode} \\
          \hline
          \hline
         \multirow{4}{*}{\textbf{R1}} & 0 & 1 $\rightarrow$ [1:1.1] & 1 & 6 $\rightarrow$ [1:1.1, 2:2] \\
         \cline{2-5}
         & 1 & 2 $\rightarrow$ [2:2] & 2 & 7 $\rightarrow$ [2:2, 3:3] \\
         \cline{2-5}
         & 2 & 3 $\rightarrow$ [3:3] & 3 & 8 $\rightarrow$ [3:3, 4:1.4] \\
         \cline{2-5}
         & 3 & 4 $\rightarrow$ [4:1.4] & 4 & NOT called \\
         \hline
         \multirow{2}{*}{\textbf{R2}} & 0 & 6 $\rightarrow$ [1:1.1, 2:2] & 6 & 9 $\rightarrow$ [1:1.1, 2:2, 3:3] \\
         \cline{2-5}
         & 2 & 3 $\rightarrow$ [3:3] & 3 & NOT called \\
         \hline
         \multirow{2}{*}{\textbf{R3}} & 0 & 5 $\rightarrow$ [2:1.1] & 5 & 10 $\rightarrow$ [2:1.1, 3:3] \\
         \cline{2-5}
         & 1 & 8 $\rightarrow$ [3:3, 4:1.4] & 8 & NOT called \\
         \hline
         \textbf{R4} & 0 & 6 $\rightarrow$ [1:1.1, 2:2] & 6 & NOT called \\
         \hline
     \end{tabular}

\vspace{-3mm}
\end{table}

\subsubsection{Comparisons with Lempel-Ziv-Welch (LZW)}
\label{sec:lzw_comparisons}
  Our prefix tree encoding algorithm is inspired by the classical compression scheme LZW. However, a key difference between LZW and our algorithm is that we preserve the row and column boundaries in the underlying tabular data, which is crucial to directly operate matrix operations on the compressed representation. For examples, our algorithm encodes each tuple separately (although the dictionary is shared) to respect the row boundaries, and the compression unit is a column\_index:value pair to
respect the column boundaries. In contrast, LZW simply encodes a blob of bytes without preserving any structure information. The reason for that is LZW was invented primarily for
string/text compression. There are other several noticeable differences between our algorithm and LZW, which are summarized in Table~\ref{tab:differences}.

\begin{table}[th!]
 \vspace{1mm}
  \caption{Differences between LZW and our prefix tree encoding. c-v stands for column-index:value.}
 \vspace{-1mm}
 \label{tab:differences}
 \centering
     \begin{tabular}{|c|c|c|}
        \hline
                  & \textbf{LZW} & \textbf{Ours} \\
        \hline
        \hline
         \textbf{Input}       & bytes & sparse encoded table \\
         \hline
         \textbf{Encode unit} & 8 bits & c-v pair \\
         \hline
         \textbf{Tree init.} & all values of 8 bits & all unique c-v pairs \\
         \hline
         \textbf{Tuple bound.} & lost & preserved \\
         \hline
         \textbf{Output} & a vector of codes & encoded table \& prefix tree first layer\\
         \hline
 \end{tabular}
 \end{table}
\vspace{1mm}


\subsection{Physical Encoding} \label{sec:physical-encoding}
The output of the logical encoding (i.e., \textbf{I} and \textbf{D} in Figure~\ref{fig:encoding-overview}) can be further
encoded physically to reduce sizes. We use two simple techniques---bit packing~\cite{lemire2015decoding} and value indexing~\cite{kourtis2008optimizing}---
that can reduce sizes without incurring significant overheads when accessing the original values.

We notice that some information in \textbf{I} and \textbf{D} can be stored using arrays of non-negative integers, and these
integers are typically small. For example, the maximal column index in \textbf{I}
of Figure~\ref{fig:encoding-overview} is 4, so 1 byte is enough to encode a single integer.
Bit packing is used to encode these arrays of small non-negative integers efficiently. Specifically, we use
$\ceil{\frac{1}{8}*\log_2^{\text{maximal\_integer+1}}}$ bytes to encode each non-negative integer in an array.
Each encoded array has a header that tells the number of integers in the array and the number of bytes
used per integer. More advanced encoding methods such as Varint~\cite{dean2009challenges} and SIMD-BP128~\cite{lemire2015decoding}, can also be used, and point to interesting directions for future work. 

Value indexing is essentially a dictionary encoding technique. That is, we store all the unique values
(excluding column indexes) in the column\_index:value pairs (e.g., \textbf{I} in Figure~\ref{fig:encoding-overview})
in an array. Then, we replace the original values with the indexes pointing to the values in the array.

Figure~\ref{fig:encoding-overview} illustrates an example of how we encode the input (e.g., \textbf{I} and \textbf{D}) to physical
bytes. For \textbf{I}, the column indexes are encoded using bit packing, while the values are encoded
using value indexing. The value indexes from applying value indexing are also encoded using bit packing.
For \textbf{D}, we concatenate the tree node indexes from all the tuples and encode them all together using bit packing.
We also encode the tuple starting indexes using bit packing.

\section{Matrix Operation Execution} \label{sec:la}
In this section, we introduce how to execute matrix operations on the TOC output. Most of the operations can directly operate on the compressed representation without decoding the original matrix.
This direct execution avoids the tedious and expensive decoding process and reduces the runtime to execute matrix operations and MGD.

Let $A$ be a TOC compressed matrix, $c$ be a scalar, $v/M$ be an uncompressed vector/matrix respectively, we discuss four common classes of matrix operations:
\begin{enumerate}
\item Sparse-safe element-wise~\cite{elgohary2016compressed} operations (e.g., $A.*c$ and $A.^2$).
\item Right multiplication operations (e.g., $A \cdot v$ and $A \cdot M$).
\item Left multiplication operations (e.g., $v \cdot A$ and $M \cdot A$).
\item Sparse-unsafe element-wise operations~\cite{elgohary2016compressed} (e.g., $A. + c$ and $A + M$).
\end{enumerate}
Informally speaking, sparse-safe operation means that zero elements in the matrix remain as zero after the operation; sparse-unsafe operation means that zero elements in the matrix may not be zero
after the operation.

\begin{figure}[th!]
\centering
    \includegraphics[width=0.97\linewidth]{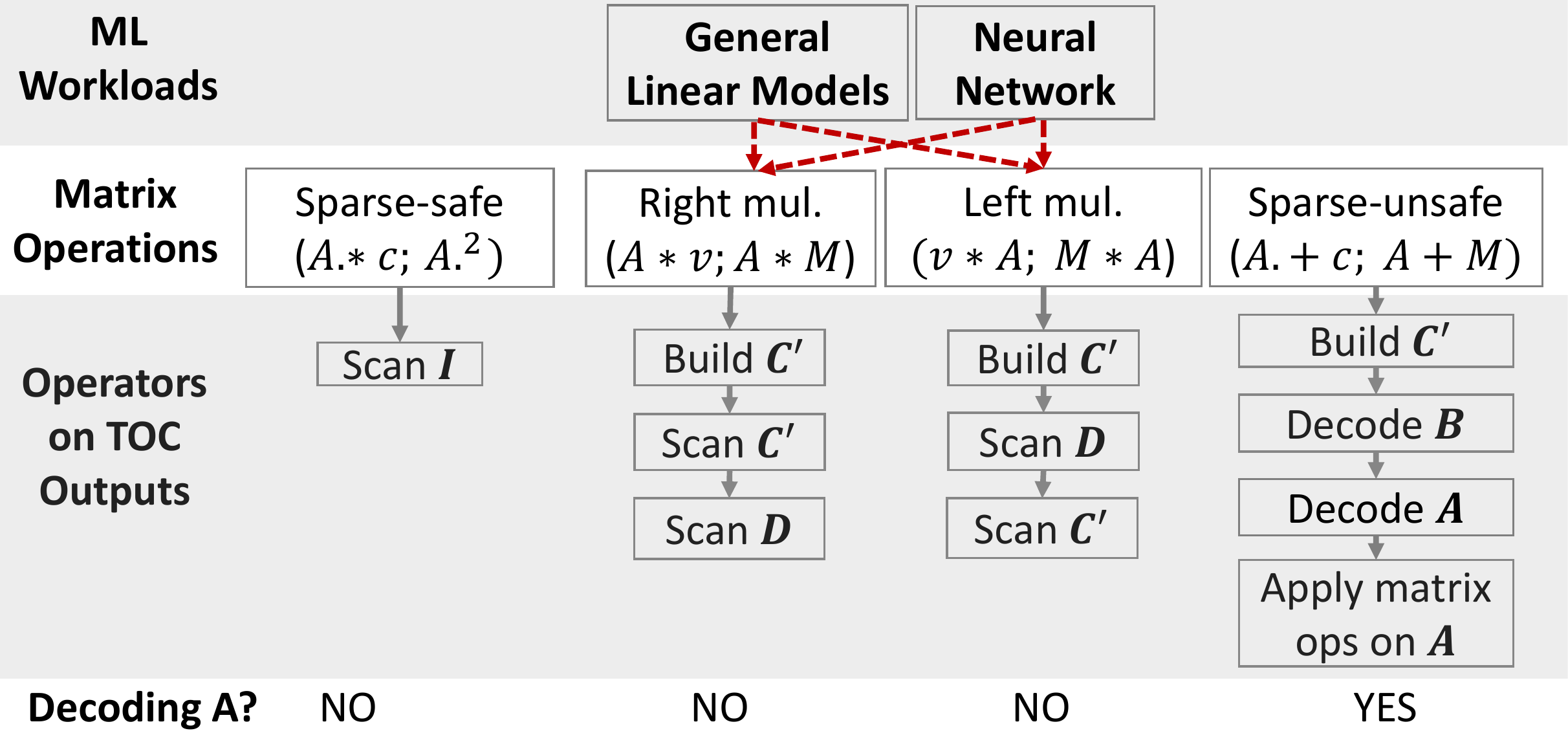}
\vspace{2mm}
    \caption{An overview of how to execute different matrix
    operations on the TOC output. For sparse-safe element-wise operations, right
    multiplication operations, and left multiplication operations, we can
    execute them on the TOC output directly. For sparse-unsafe element-wise operations,
    we need to fully decode the input $A$ and then apply the operation on $A$.}
\label{fig:decoding_overview}
\vspace{-3mm}
\end{figure}

Figure~\ref{fig:decoding_overview} gives an overview of how to execute different matrix operations on the TOC output.
The first three classes of operations can directly operate over the compressed representation without decoding
the original matrix. The last class of operations needs to decode the original matrix. However, it is less likely
to be used for training machine learning models because it changes the input data.

\subsection{Shared Operators}
In this subsection, we discuss some shared operators for executing matrix operations on the TOC output.
\subsubsection{Access Values of I and D From Physical Bytes}
As shown in Figure~\ref{fig:decoding_overview}, executing many matrix operations requires scanning \textbf{I} or \textbf{D}, which are encoded to physical bytes using bit packing and value indexing as explained in \S~\ref{sec:physical-encoding}.
Thus, we briefly discuss how to access values of \textbf{I} and \textbf{D} from encoded physical bytes.

To access a non-negative integer encoded using bit packing, one
can simply seek to the starting position of the integer, and cast its bytes to uint\_8, uint\_16, or 
uint\_32 respectively. Unfortunately, most programming languages do not support uint\_24 natively. 
Nevertheless, one can copy the bytes into an uint\_32 and mask its leading byte as zero.

To access values encoded using value indexing, one can look up the array which stores the unique values using the value indexes.

\subsubsection{Build Prefix Tree For Decoding} \label{sec:build_prefix_tree}
As shown in Figure~\ref{fig:decoding_overview}, executing all matrix operations except for
sparse-safe element-wise operations needs to build the prefix tree $\mathbf{C'}$, which is a simplified
variant of the prefix tree \textbf{C} built during encoding.
Each node in $\mathbf{C'}$ has the same index and key with the node in 
\textbf{C}. The difference is that each node in $\mathbf{C'}$ stores the index to its parent, but it does NOT store
indexes to its children. Table~\ref{tab:c_prime_example} demonstrates an example of $\mathbf{C'}$.

\begin{table}[th!]
     \vspace{1mm}
    \caption{An example of $\mathbf{C'}$, which is a simplified variant of \textbf{C} in Figure~\ref{fig:encoding-overview}.
    Each node in $\mathbf{C'}$ only stores the index to its parent but NOT
    indexes to its children.}
\label{tab:c_prime_example}
 \centering
       \begin{tabular}{|c|c|c|c|c|c|c|c|c|c|c|c|}
         \hline
         \small{\textbf{Index}} & 0 & 1 & 2 & 3 & 4 & 5 & 6 & 7 & 8 & 9 & 10 \\
         \hline
         \small{\textbf{Key}} & & 1:1.1 & 2:2 & 3:3 & 4:1.4 & 2:1.1 & 2:2 & 3:3 & 4:1.4 & 3:3 & 3:3 \\
         \hline
         \small{\textbf{ParentIndex}} & & 0 & 0 & 0 & 0 & 0 & 1 & 2 & 3 & 6 & 5 \\
         \hline
     \end{tabular}

 \end{table}

Algorithm~\ref{alg:build_prefix_tree_c_prime} presents how to build $\mathbf{C'}$. There are two main phases in 
Algorithm~\ref{alg:build_prefix_tree_c_prime}. In phase \rom{1}, $\mathbf{C'}$ and \textbf{F} are both initialized by
\textbf{I}, where \textbf{F} stores the first column\_index:value pair of the sequence represented by each tree node.

In phase \rom{2}, we scan \textbf{D} to build $\mathbf{C'}$ mimicking how \textbf{C} is built in 
Algorithm~\ref{alg:prefix_tree_encoding}. From line 11 to line 13 of
Algorithm~\ref{alg:build_prefix_tree_c_prime}, we add a new prefix tree node indexed by idx\_seq\_num. More specifically,
the new tree node is a child of the tree node indexed by \textbf{D}[$i$][$j$] (line 11), the first column\_index:value 
pair of the sequence represented by the new tree node is the same with its parent (line 12), and the key of the new tree node
is the first column\_index:value pair of the sequence represented by the next tree node indexed by \textbf{D}[$i$][$j+1$] (line 13).

\begin{algorithm} [th!]
\caption{Build Prefix Tree $\mathbf{C'}$} \label{alg:build_prefix_tree_c_prime}
\begin{algorithmic}[1]
\Function{BuildPrefixTree}{\textbf{I}, \textbf{D}}
	\State \textbf{inputs:} column\_index:value pairs in the first layer
    \Statex of the prefix tree \textbf{I} and encoded table \textbf{D}
    \State \textbf{outputs:} A prefix tree used for decoding $\mathbf{C'}$
    \For {$i \gets$ 1 to len(\textbf{I})} \Comment{phase \rom{1}: initialize with \textbf{I}}
    	\State $\mathbf{C'}[i]$.key $\gets \textbf{I}[i-1]$
        \State $\mathbf{C'}[i]$.parent $\gets 0$
        \State $\textbf{F}{[i]} \gets \textbf{I}[i-1]$
        \Comment{\textbf{F} stores the first column\_index:value pair of the sequence of the node}

    \EndFor
    \State idx\_seq\_num $\gets$ len(\textbf{I}) + 1
    \For {$i \gets $ 0 to len(\textbf{D}) - 1} \Comment{phase \rom{2}: build $\mathbf{C'}$}
        \For {$ j \gets $ 0 to len(\textbf{D}[$i$]) -2 } \Comment{skip last element}
        	\State $\mathbf{C'}$[idx\_seq\_num].parent $\gets$ \textbf{D}[$i$][$j$]
            \State \textbf{F}[idx\_seq\_num] $\gets$ \textbf{F}[\textbf{D}[$i$][$j$]]
            \State $\mathbf{C'}$[idx\_seq\_num].key $\gets$ \textbf{F}[\textbf{D}[$i$][$j+1$]]
            \State idx\_seq\_num $\gets$ idx\_seq\_num + 1
        \EndFor 
    \EndFor
    \State \textbf{return ($\mathbf{C'}$)}
\EndFunction
\end{algorithmic}
\end{algorithm}

\subsection{Sparse-safe Element-wise Operations}
To execute sparse-safe element-wise operations (e.g., $A.* c$ and $A.^2$) on the TOC output directly, one can simply scan and modify
\textbf{I} because all the unique column\_index:value pairs in $A$ are stored in \textbf{I}. Algorithm
~\ref{alg:matrix_times_scalar} demonstrates how to execute matrix times scalar operation (i.e., $A. * c$) on the TOC output.
Algorithms for other sparse-safe element-wise operations are similar.

\begin{algorithm} [th!]
\caption{Execute matrix times scalar operation (i.e., $A. * c$) on the TOC output.} \label{alg:matrix_times_scalar}
\begin{algorithmic}[1]
\Function{MatrixTimesScalar}{\textbf{I}, c}
	\State \textbf{inputs:} column\_index:value pairs in the first layer
    \Statex of the prefix tree \textbf{I} and a scalar c
    \State \textbf{outputs:} the modified \textbf{I}
	\For {$i \gets $ 0 to len(\textbf{I}) -1}
    	\State $\textbf{I}[i]$.value $\gets$ $\textbf{I}[i]$.value * c
    \EndFor
    \State \textbf{return(I)}
\EndFunction
\end{algorithmic}
\end{algorithm}

\subsection{Right Multiplication Operations}
\label{right_multiplication_vector}
We first do some mathematical analysis to transform the uncompressed execution of right multiplication operations to the compressed execution that operates directly on the TOC output without decoding the original table. The analysis also proves the algorithm correctness since the algorithm follows the transformed form directly. Then, we demonstrate the detailed algorithm. In the rest of this subsection, we use $A \cdot v$ as an example. We put the result of $A \cdot M$ (similar to $A \cdot v$) in Appendix~\ref{more_algorithms} for brevity.

\begin{theorem}\label{theorem:matrix_times_vector}
Let $A \in \Re^{n \times m}$, $v \in \Re^{m \times 1}$, \textbf{D} be the output of TOC on $A$,
$\mathbf{C'}$ be the prefix tree built for decoding, $\mathbf{C'}[i].seq$ be the sequence of the tree node defined in
\S~\ref{sec:api}, $\mathbf{C'}[i].key$ be the key of the tree node defined in \S~\ref{sec:build_prefix_tree},
and $\mathbf{C'}[i].parent$ be the parent index of the tree node defined in \S~\ref{sec:build_prefix_tree}.
Note that $\mathbf{C'}[i].key$ and $\mathbf{C'}[i].seq$ are both sparse representations of vectors
(i.e., $\mathbf{C'}[i].key \in \Re^{1 \times m}$ and $\mathbf{C'}[i].seq \in \Re^{1 \times m}$).
Define function $\mathcal{F}(x): \aleph \rightarrow \Re$ to be

\begin{align} \label{equation:matrix_times_vector_definition}
	\mathcal{F}(x) = \mathbf{C'}[x].seq \cdot v, x = 1, 2, ..., \mathrm{len}(\mathbf{C'})-1.
\end{align}
Then, we have
\begin{align} \label{equation:matrix_times_vector_first_equation}
A[i, :] \cdot v = & \sum_{j=0}^{\mathrm{len}(\mathbf{D}[i, :])-1} \mathcal{F}(\mathbf{D}[i][j]), i = 0, 1, ..., n-1
\end{align}
and
\begin{align}
	\mathbf{C'}[i].seq =&~\mathbf{C'}[i].key + \mathbf{C'}[\mathbf{C'}[i].parent].seq, & \nonumber \\
                                                                   & i = 1, 2, ..., \mathrm{len}(\mathbf{C'})-1.
    \label{equation:vector_times_matrix_second_equation}
\end{align}
\end{theorem}

\begin{proof}
See Appendix~\ref{appendix:matrix_times_vector}
\end{proof}

\noindent \textbf{\em Remark on Theorem~\ref{theorem:matrix_times_vector}}. $A \cdot v$ can be directly executed on the TOC output following 
Equation~\ref{equation:matrix_times_vector_first_equation} by scanning $\mathbf{C'}$ first and scanning $\mathbf{D}$ second. 
The detailed steps are demonstrated in Algorithm~\ref{alg:matrix_times_vector}.

First, we scan $\mathbf{C'}$ to compute $\mathcal{F}$ function defined in Equation~\ref{equation:matrix_times_vector_definition}
(lines 5-7 in Algorithm~\ref{alg:matrix_times_vector}). The dynamic programming technique is used following
Equation~\ref{equation:vector_times_matrix_second_equation}. Specifically, we use \textbf{H}[$i$] to remember the computed
value of $\mathcal{F}(i)$. We compute each
\textbf{H}[$i$] as the sum of $\mathbf{C'}[i].key \cdot v$ and \textbf{H}[$\mathbf{C'}$[$i$].parent], which is computed already.

Second, we scan \textbf{D} to compute $A \cdot v$ and store it in \textbf{R} following Equation~\ref{equation:matrix_times_vector_first_equation}
(lines 8-11 in Algorithm~\ref{alg:matrix_times_vector}). For each \textbf{D}[$i$][$j$], we simply add 
\textbf{H}[\textbf{D}[$i$][$j$]] to \textbf{R}[$i$].

\begin{algorithm} [th!]
\caption{Execute matrix times vector operation (i.e., $A \cdot v$) on the TOC output.} \label{alg:matrix_times_vector}
\begin{algorithmic}[1]
\Function{MatrixTimesVector}{\textbf{D}, \textbf{I}, $v$}
	\State \textbf{inputs:} column\_index:value pairs in the first layer of
    the prefix tree \textbf{I}, encoded table \textbf{D}, and vector $v$
    \State \textbf{outputs:} the result of $A \cdot v$ in \textbf{R}
    \State $\mathbf{C'} \gets$ \Call{BuildPrefixTree}{\textbf{I}, \textbf{D}}
    \State \textbf{H} $\gets \overrightarrow{0}$ \Comment{initialize as a zero vector}
    \For {$i$ = 1 to len($\mathbf{C'}$)-1} \Comment{scan $\mathbf{C'}$ to compute \textbf{H}}
    	\State \textbf{H}[$i$] $ \gets \mathbf{C'}$[$i$].key $\cdot v$ + \textbf{H}[$\mathbf{C'}$[$i$].parent]
    \EndFor
    \State \textbf{R} $\gets \overrightarrow{0}$ \Comment{initialize as a zero vector}
    \For {$i$ = 0 to len(\textbf{D})-1} \Comment{scan \textbf{D} to compute \textbf{R}}
    	\For {$j$ = 0 to len(\textbf{D}[$i$,:])-1}
        	\State \textbf{R}[$i$] $\gets$ \textbf{R}[$i$] + \textbf{H}[\textbf{D}[$i$][$j$]]
        \EndFor
    \EndFor
    \State \textbf{return(R)}
\EndFunction
\end{algorithmic}
\end{algorithm}

\subsection{Left Multiplication Operations}
\label{left_multiplication_vector}
We first give the mathematical analysis and then present the detailed algorithm. The reason for doing so is similar to that is given in \S~\ref{right_multiplication_vector}. We only demonstrate the result of $v \cdot A$ and put the result of $M \cdot A$ to Appendix~\ref{more_algorithms} for brevity.

\begin{theorem} \label{theorem:vector_times_matrix}
Let $A \in \Re^{n \times m}$, $v \in \Re^{1 \times n}$, \textbf{D} be the output of TOC on $A$,
$\mathbf{C'}$ be the prefix tree built for decoding, $\mathbf{C'}[i]$.seq be the sequence of the tree node defined in 
\S~\ref{sec:api}, $\mathbf{C'}[i].key$ be the key of the tree node defined in \S~\ref{sec:build_prefix_tree},
and $\mathbf{C'}[i].parent$ be the parent index of the tree node defined in \S~\ref{sec:build_prefix_tree}.
Note that $\mathbf{C'}[i].key$ and $\mathbf{C'}[i].seq$ are both sparse representations of vectors
(i.e., $\mathbf{C'}[i].key \in \Re^{1 \times m}$ and $\mathbf{C'}[i].seq \in \Re^{1 \times m}$).
Define function $\mathcal{G}(x): \aleph \rightarrow \Re$ to be

\begin{align}
	\mathcal{G}(x) = \sum_{\substack{\mathbf{D}[i, j]=x,
    					\forall i \in \aleph,
                        \forall j \in \aleph}}
                        v[i],~x = 1, 2, ..., \mathrm{len}(\mathbf{C'})-1. \label{equation:vector_times_matrix_definition}
\end{align}
Then, we have

\begin{align}
	v \cdot A =& \sum_{i=1}^{\mathrm{len}(\mathbf{C'})-1} \mathbf{C'}[i].seq \cdot \mathcal{G}(i).
    \label{equation:vector_times_matrix_first_equation}
\end{align}
\end{theorem}

\begin{proof}
See Appendix~\ref{appendix:vector_times_matrix}.
\end{proof}

\noindent \textbf{\em Remark on Theorem~\ref{theorem:vector_times_matrix}}. We can compute $v \cdot A$ following 
Equation~\ref{equation:vector_times_matrix_first_equation} by simply scanning \textbf{D} first and scanning $\mathbf{C'}$ second. Algorithm~\ref{alg:vector_times_matrix} presents the detailed steps.
First, we scan \textbf{D} to compute function $\mathcal{G}$ 
defined in Equation~\ref{equation:vector_times_matrix_definition}. Specifically, we initialize $\mathbf{H}$ as a zero vector,
and then add $v[i]$ to $\mathbf{H}[\mathbf{D}[i][j]]$ for each $\mathbf{D}[i][j]$
(lines 6-8 in Algorithm~\ref{alg:vector_times_matrix}). After this step is done, $\mathcal{G}(i) = \mathbf{H}[i], i$ = 1, 2, $\dots$, len$(\mathbf{C'})-1$.

Second, we scan $\mathbf{C'}$ backwards to actually compute $v \cdot A$ and store it in \textbf{R} following Equation~\ref{equation:vector_times_matrix_first_equation} (lines 10-12 in Algorithm~\ref{alg:vector_times_matrix}). The dynamic
programming technique is used following Equation~\ref{equation:vector_times_matrix_second_equation}. Specifically,
for each $\mathbf{C'}[i]$, we add $\mathbf{C'}[i].key \cdot \mathbf{H}[i]$ to \textbf{R} and add $\mathbf{H}[i]$ to
$\mathbf{H}[\mathbf{C'}[i].parent]$.

\begin{algorithm} [th!]
\caption{Execute vector times matrix operation (i.e., $v \cdot A$) on the TOC output.} \label{alg:vector_times_matrix}
\begin{algorithmic}[1]
\Function{VectorTimesMatrix}{\textbf{D}, \textbf{I}, $v$}
	\State \textbf{inputs:} column\_index:value pairs in the first layer of
    the prefix tree \textbf{I}, encoded table \textbf{D}, and vector $v$
    \State \textbf{outputs:} the result of $v \cdot A$ in \textbf{R}
    \State $\mathbf{C'} \gets$ \Call{BuildPrefixTree}{\textbf{I}, \textbf{D}}
    \State \textbf{H} $\gets \overrightarrow{0}$ \Comment{initialize as a zero vector}
    \For {$i$ = 0 to len(\textbf{D})-1} \Comment{scan \textbf{D} to compute \textbf{H}}
    	\For {$j$ = 0 to len(\textbf{D}[$i$,:]) -1}
        	\State \textbf{H}[\textbf{D}[$i$][$j$]] $\gets$ $v$[$i$] + \textbf{H}[\textbf{D}[$i$][$j$]]
        \EndFor
    \EndFor
    \State \textbf{R} $\gets \overrightarrow{0}$ \Comment{initialize as a zero vector}
    \For {$i$ = len($\mathbf{C'}$) -1 to 1} \Comment{scan $\mathbf{C'}$ to compute \textbf{R}}
    	\State \textbf{R} $\gets$ \textbf{R} + $\mathbf{C'}$[$i$].key $\cdot$ \textbf{H}[$i$]
        \State \textbf{H}[$\mathbf{C'}[i]$.parent] $\gets$ \textbf{H}[$\mathbf{C'}$[i].parent] + \textbf{H}[$i$]
    \EndFor
    \State \textbf{return(R)}
\EndFunction
\end{algorithmic}
\end{algorithm}

\subsection{Sparse-unsafe Element-wise Operations}
For sparse-unsafe element-wise operations (e.g., $A.+c$ and $A+M$), we need to fully decode $A$ first and then
execute the operations on $A$. Although this process is slow due to the tedious
decoding step, fortunately, sparse-unsafe element-wise operations are less likely to be used for training ML models because the input data is changed. Algorithm~\ref{alg:matrix_plus_scalar} presents the detailed steps.

\begin{algorithm} [th!]
\caption{Execute matrix plus scalar element-wise operation (i.e., $A.+c$) on the TOC output.}
\label{alg:matrix_plus_scalar}
\begin{algorithmic}[1]
\Function{MatrixPlusScalar}{\textbf{D}, \textbf{I}, $c$}
	\State \textbf{inputs:} column\_index:value pairs in the first layer of
    the prefix tree \textbf{I}, encoded table \textbf{D}, and scalar $c$
    \State \textbf{outputs:} the result of $A.+c$ in \textbf{R}
    \State $\mathbf{C'} \gets$ \Call{BuildPrefixTree}{\textbf{I}, \textbf{D}}
    \For {$i$ = 0 to len(\textbf{D})-1}
    	\State \textbf{B}[$i$] $\gets$ [] \Comment{initialize \textbf{B}[$i$] as an empty vector}
        \For {$j$ = 0 to len(\textbf{D}[$i$,:])-1}
        	\State reverse\_seq $\gets$ []
            \State tree\_index $\gets$ \textbf{D}[$i$][$j$]
            \While{tree\_index $\neq$ 0} \Comment{backtrack $\mathbf{C'}$ to get the reversed sequence of the tree node \textbf{D}[$i$][$j$]}
            	\State reverse\_seq.Append($\mathbf{C'}$[tree\_index].key)
                \State tree\_index $\gets \mathbf{C'}$[tree\_index].parent
            \EndWhile
            \For {$k$ = len(reverse\_seq)-1 to 0}
            	\State \textbf{B[$i$]}.Append(reverse\_seq[$k$])
            \EndFor
        \EndFor
    \EndFor
    \State $A \gets$ TransformToDense(\textbf{B})
    \State \textbf{R} $\gets A.+c$
    \State \textbf{return(R)}
\EndFunction
\end{algorithmic}
\end{algorithm}

\subsection{Time Complexity Analysis}
%
We give detailed time complexity analysis of different matrix operations except for $A \cdot M$ and $M \cdot A$, which we put to Appendix~\ref{more_complexity_analysis} for brevity.
For $A.*c$, we only need to scan \textbf{I}, so the time complexity is $\mathcal{O}(|\mathbf{I}|)$.

For $A \cdot v$ and $v \cdot A$, we need to build $\mathbf{C'}$ , scan $\mathbf{C'}$, and scan \textbf{D}. As shown in Algorithm~\ref{alg:build_prefix_tree_c_prime}, building $\mathbf{C'}$
needs to scan \textbf{I} and \textbf{D}, and $|\mathbf{C'}| = |\mathbf{I}| + |\mathbf{D}|$. So the complexity of building and scanning $\mathbf{C'}$ are $\mathcal{O}(|\mathbf{I}| + |\mathbf{D}|)$.
Overall, the complexity of $A \cdot v$ and $v \cdot A$ are $\mathcal{O}(|\mathbf{I}| + |\mathbf{D}|)$. This indicates that the computational redundancy incurred by the data redundancy
is generally avoided by TOC matrix execution algorithms. Thus, theoretically speaking, TOC matrix execution algorithms have good performance when there are many data redundancies.

For $A.+c$, we need to decompress \textbf{A} first. Similar to LZW, the decompression of TOC is linear in the sense that each element has to be outputted and the cost of
each output element is constant. Thus, the complexity of decompressing \textbf{A} is $\mathcal{O}(|\mathbf{A}|)$. Overall, the complexity of $A.+c$ is also $\mathcal{O}(|\mathbf{A}|)$.

\section{Experiments} \label{sec:expts}
In this section, we answer the following questions:
\begin{enumerate}
    \item Can TOC compress mini-batches effectively?
    \item Can common matrix operations be executed efficiently on TOC compressed mini-batches?
    \item Can TOC reduce the end-to-end MGD runtimes significantly for training common machine learning models?
    \item Can TOC compress/decompress mini-batches fastly?
\end{enumerate}

\vspace{1mm}
\noindent \textbf{Summary of Answers.} We answer these questions positively by conducting extensive experiments. First,  on datasets with moderate sparsity, TOC reduces mini-batch sizes notably with compression ratios up to 51x.
Compression ratios of TOC are up to 3.8x larger than
the state-of-the-art light-weight matrix compression schemes, and comparable to general compression schemes such as Gzip.
Second, the matrix operation runtime of TOC is comparable to the light-weight matrix compression schemes, and up to 20,000x
better than the state-of-the-art general compression schemes. Third, TOC reduces
the end-to-end MGD runtimes by up to 1.4x, 5.6x, and 4.8x compared to the state-of-the-art compression schemes for
training Neural network, Logistic regression, and Support vector machine, respectively. TOC also reduces the MGD runtimes by up to 10.2x compared to
the best encoding methods used in popular machine learning systems: Bismarck, ScikitLearn, and TensorFlow.
Finally, the compression speed of TOC is much faster than Gzip but slower than Snappy, whereas the decompression speed of TOC is faster than both Gzip and Snappy.

\vspace{1mm}
\noindent \textbf{Datasets.}
We use six real-world datasets. The first four datasets we chose have moderate sparsity, which is a typical phenomenon for enterprise machine learning~\cite{ashari2015optimizing,harnik2012estimation}. Rcv1 and Deep1Billion represent the extremely sparse and dense dataset respectively.
Table~\ref{tab:dataset_statistics} lists the dataset statistics.
\begin{table}[!htbp]
 \vspace{1mm}
 \caption{Dataset statistics. Except for Deep1Billion, which is in the binary format, we report the sizes of the datasets stored in the text format. Sparsity is defined as $\frac{\# \text{ of non zero values}} {\# \text{ of total values}}$.}
 \vspace{-1mm}
 \label{tab:dataset_statistics}
 \centering
     \begin{tabular}{|c||c|c|c|}
         \hline
	     \textbf{Dataset} & \textbf{Dimensions} & \textbf{Size} & \textbf{Sparsity} \\
         \hline
 		 \hline 
	     US Census~\cite{elgohary2016compressed} & 2.5 M * 68 & 0.46 GB & 0.43 \\
         \hline
	     ImageNet~\cite{elgohary2016compressed} &  1.2 M * 900 & 2.8 GB & 0.31 \\
         \hline
	     Mnist8m~\cite{elgohary2016compressed} & 8.1 M * 784 & 11.3 GB & 0.25 \\
         \hline
	     Kdd99~\cite{Lichman:2013} & 4 M * 42 & 1.6 GB & 0.39 \\
         \hline
       Rcv1~\cite{amini2009learning} & 800 K * 47236 & 0.96 GB & 0.0016  \\
       \hline
       Deep1Billion~\cite{babenko2016efficient} & 1 B * 96 & 475 GB &  1.0 \\
         \hline
 \end{tabular}

 \end{table}

\noindent \textbf{Compared Methods.}
We compare TOC with one baseline (DEN), four light-weight matrix compression schemes (CSR, CVI, DVI, and CLA),
and two general compression schemes (Snappy and Gzip). A brief summary of these methods is as follows:
\begin{enumerate}
  \item DEN: This is the standard dense binary format for dense matrices.
    We store the matrix row by row and each value is encoded using IEEE-754 double format.
    Categorical features are encoded using the standard one-hot (dummy) encoding~\cite{garavaglia1998smart}.
  \item CSR: This is the standard compressed sparse row encoding for sparse matrices. We store the matrix row by row.
    For each row, we only store the non-zero values and associated column indexes.
  \item CVI: This format is also called as CSR-VI~\cite{kourtis2008optimizing,elgohary2016compressed}. We first encode the matrix
    using CSR and then encode non-zero values via the value indexing in Section~\ref{sec:physical-encoding}.
  \item DVI: We first encode the matrix using DEN and then encode the values via the value indexing in Section~\ref{sec:physical-encoding}.
  \item CLA: This method~\cite{elgohary2016compressed} divides the matrix into different column-groups and compresses each column-group column-wisely. Note
  that matrix operations can be executed on compressed CLA matrices directly.
  \item Snappy: We compress the serialized bytes of DEN using Snappy.
  \item Gzip: We use Gzip to compress the serialized bytes of DEN.
\end{enumerate}

\vspace{1mm}
\noindent \textbf{Machine and System Setup.}
All experiments were run on Google cloud \footnote{https://cloud.google.com/} using a typical machine with a $4$ core, $2.2$ GHz Intel Xeon CPU, $15$GB RAM (unless otherwise specified), and OS Ubuntu $14.04.1$.
We did not choose a more powerful machine because of the higher cost. For example, our machine
costs \$$131$ per month, while a machine with $4$ cores and $180$ GB RAM costs \$$912$ per month. Thus, our techniques can save costs for ML workloads, especially in such cloud settings.

Our techniques were implemented in C++ and compiled using g++ $4.8.4$ with the flag O3 optimization.
We also compare with four machine learning systems: ScikitLearn 0.19.1\footnote{http://scikit-learn.org/stable/}, Systemml 1.3.0\footnote{https://systemml.apache.org/}, Bismarck 2.0\footnote{http://i.stanford.edu/hazy/victor/bismarck/}, and TensorFlow\footnote{https://www.tensorflow.org/}. Furthermore, we integrate TOC into Bismarck to realize fair comparison. We put the integration detail into Appendix~\ref{appendix:integration} for brevity.

\subsection{Compression Ratios} \label{compression_ratios}
\vspace{1mm}
\begin{figure*}[!htbp]
\centering
    \includegraphics[width=0.97\linewidth]{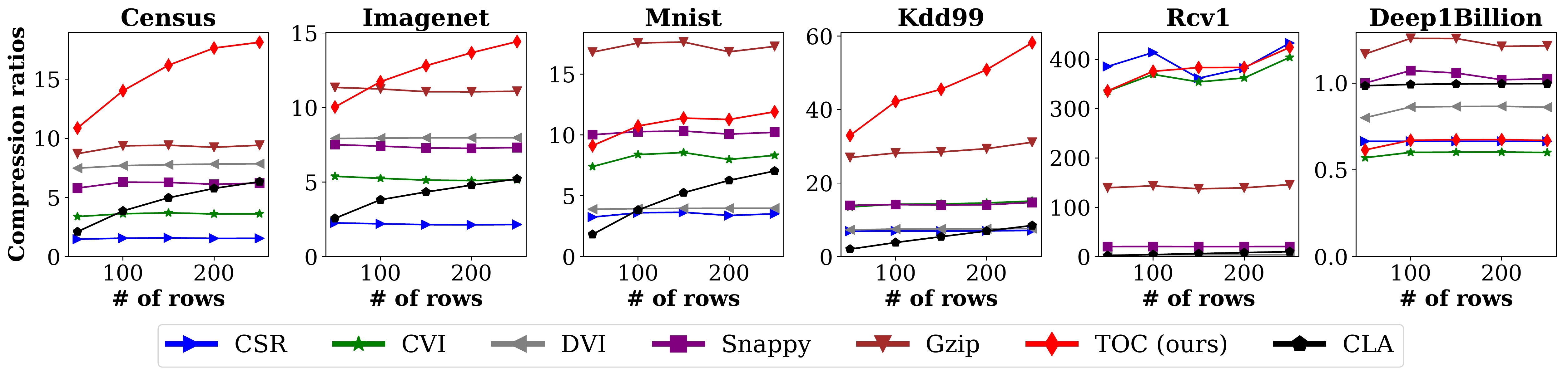}
\vspace{-2mm}
    \caption{Compression ratios of different methods on mini-batches with varying sizes.}
\label{fig:compression_ratios}
\vspace{-3mm}
\end{figure*}

\begin{figure*}[!htbp]
\centering
    \includegraphics[width=0.97\linewidth]{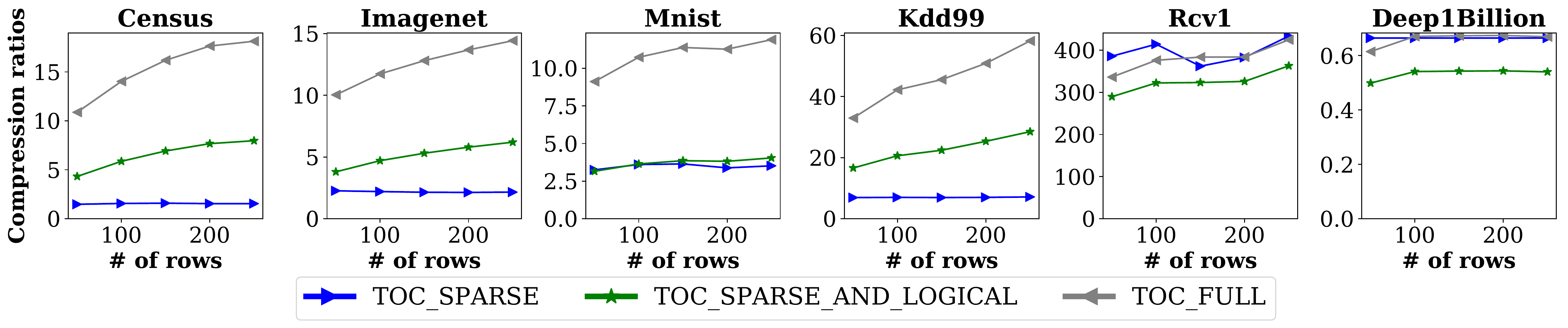}
    \caption{Compression ratios of TOC variants on varying size mini-batches. TOC\_SPARSE uses sparse encoding. TOC\_SPARSE\_AND\_LOGICAL uses sparse and logical encoding. TOC\_FULL uses all the encoding techniques.}
\label{fig:compression_ratio_ablation_study}
\vspace{-3mm}
\end{figure*}

\begin{figure*}[!htbp]
\centering
    \includegraphics[width=0.97\linewidth]{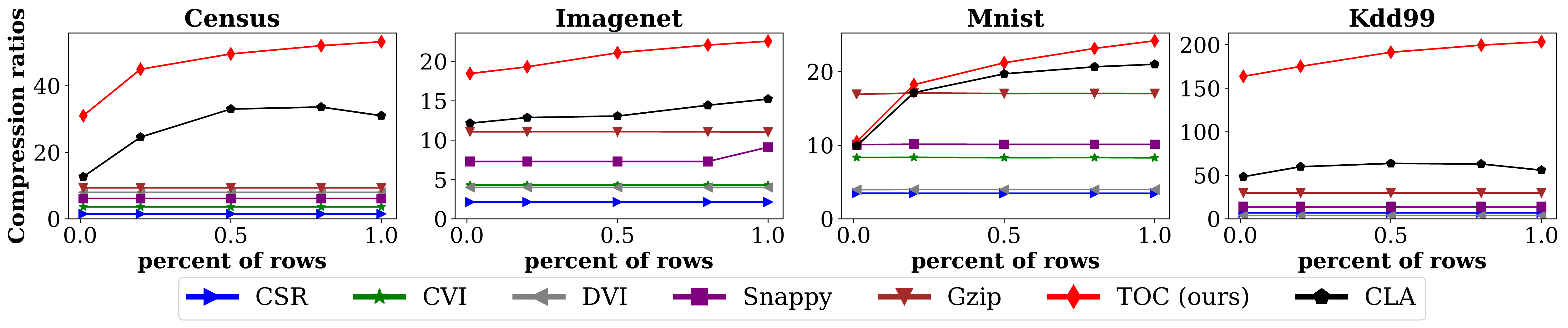}
    \caption{Compression ratios of different methods on large mini-batches. The x-axis is the percent of rows of the whole dataset in the mini-batch.}
\label{fig:compression_ratios_percent}
\vspace{-3mm}
\end{figure*}

\noindent \textbf{Setup.}
We are not aware of a first-principle way in literature to set mini-batch sizes (\# of rows in a mini-batch). In practice, the mini-batch 
size typically depends on system constraints (e.g. number of CPUs) and is set to some number ranging from
10 to 250~\cite{mishkin2016systematic}. Thus, we tested five mini-batch sizes 50, 100, 150, 200, and 250,
which cover the most common use cases. Compression ratio is defined as the uncompressed mini-batch size (encoded using 
DEN) over the compressed mini-batch size. We implemented DEN, CSR, CVI, and DVI by ourselves but use CLA from Systemml and Gzip/Snappy from standard libraries. We tested mini-batches from all the real datasets with the sizes mentioned above. 

\vspace{1mm}
\noindent \textbf{Overall Results.} Figure~\ref{fig:compression_ratios}
presents the overall results. For the very sparse dataset Rcv1, CSR is the best encoding method and TOC's performance is similar to CSR.
For the very dense dataset Deep1Billion, which does not contain repeated subsequences of column values, Gzip is the best method but it only
achieved a marginal 1.15x compression ratio. CSR and TOC have similar performance because of the sparse encoding.

For the other 4 datasets with moderate sparsity, TOC has larger compression ratios than all the other methods except on Mnist,
where TOC is inferior to Gzip. The main reason is that Mnist does not contain many repeated subsequences of column values that TOC
logical encoding can exploit, this is also verified by the ablation study in Figure~\ref{fig:compression_ratio_ablation_study}.

Overall, TOC is suitable for datasets of moderate sparsity, which are commonly used datasets in enterprise ML. TOC is not suitable for very sparse datasets
and very dense datasets that do not contain repeated subsequences of columns values. Note that these datasets are challenging for other compression methods too.
Nevertheless, one can simply test TOC on a mini-batch sample and figure out if TOC is suitable for the dataset.

\noindent \textbf{Ablation Study.} We conduct an ablation study to show the effectiveness of different components (e.g., sparse encoding,
logical encoding, and physical encoding) in TOC. TOC\_SPARSE\_AND\_LOGICAL compresses better than TOC\_SPARSE. TOC\_FULL with all the encoding techniques compresses the best.
This shows the effectiveness of all our encoding components. Figure~\ref{fig:compression_ratio_ablation_study} shows the ablation study results.

\noindent \textbf{Large Mini-batches.} We compare different compression methods on large mini-batches. 
Figure~\ref{fig:compression_ratios_percent} shows the results. As the mini-batch size grows, TOC becomes more competitive.
When the percent of rows of the whole dataset in the mini-batch is 1.0, this is essentially batch gradient descent (BGD)
and TOC has the best compression ratio in this case. This shows the potential of applying TOC to BGD related workloads.

\subsection{Matrix Operation Runtimes} \label{matrix_operations}
\vspace{1mm}
\begin{figure*}[th!]
\centering
    \includegraphics[width=0.97\linewidth]{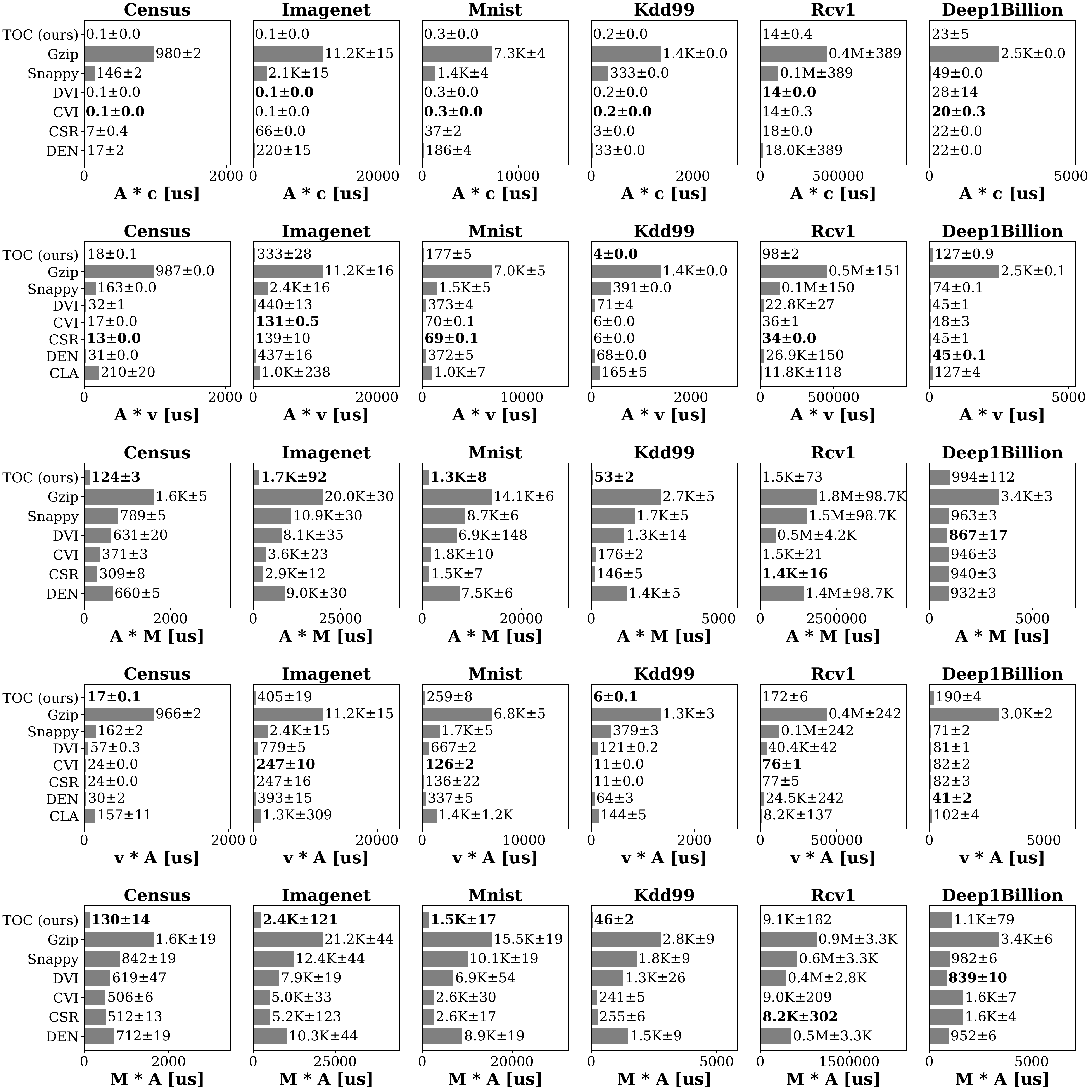}
\vspace{-2mm}
    \caption{Average runtimes (5 runs) and 95\% confidence intervals to execute different matrix operations on compressed mini-batches. From top to bottom
    are different matrix operations, where $c$ is a scalar, $v$ is an
    uncompressed vector, $M$ is an uncompressed matrix, and $A$ is the compressed matrix. From left to right are different datasets.}
\label{fig:matrix_operation_runtimes}
\end{figure*}

\noindent \textbf{Setup.}
We tested three classes of matrix operations: sparse-safe element-wise operation ($A \cdot c$), left 
multiplication operations ($v \cdot A$ and $M \cdot A$), and right multiplication operations ($A \cdot v$ and $A \cdot M$), where
$c$ is a scalar, $v$ is an uncompressed vector, $M$ is an uncompressed matrix, and $A$ is the compressed mini-batch. We set mini-batch size as 250 (results for other mini-batch sizes are similar). Figure~\ref{fig:matrix_operation_runtimes} presents the results.

\vspace{1mm}
\noindent \textbf{Sparse-safe Element-wise Operations ($A \cdot c$).} In general, DVI, CVI, and TOC are fastest methods. This shows the effectiveness of value indexing~\cite{kourtis2008optimizing} which is used by all these methods. It is noteworthy that TOC can be four orders of magnitude faster than Gzip and Snappy (e.g. on Imagenet). The slowness of these general compression schemes is caused by their significant decompression overheads.

\vspace{1mm}
\noindent \textbf{Right Multiplication Operations ($A \cdot v$ and $A \cdot M$).}
For $A \cdot v$, CSR/DEN are the best methods for Rcv1/Deep1Billion due to their extreme sparsity/density respectively.
For the remaining datasets of moderate sparsity, DEN, CSR, CVI, DVI, and TOC
are fastest methods. CLA and general compression schemes like Snappy and Gzip are much slower.  We do see that TOC
is 2-3x slower than CSR on dataset Imagenet and Mnist. There are two main reasons for the slowness. First, building the prefix tree
$\mathbf{C'}$ in TOC takes extra time. Second, TOC compression ratios over CSR compression ratios are
relatively small on these datasets, which render the computational redundancies exploited by TOC on these datasets also smaller.

For $A \cdot M$, we set the row size of $M$ as 20. TOC is consistently the fastest on
all the datasets except for Rcv1 and Deep1Billion due to its extreme sparsity/density. CLA in Systemml does not
support $A \cdot M$ yet, thus CLA is excluded.

\vspace{1mm}
\noindent \textbf{Left Multiplication Operations ($v \cdot A$ and $M \cdot A$).} The results of left multiplication operations are similar to right multiplication operations. Thus, we leave them for brevity.

\vspace{1mm}
\noindent \textbf{Summary.}
Overall, TOC achieves the best runtime performance on operations: $A \cdot c$, $A \cdot M$, and $M \cdot A$. TOC can be 2-3x slower than the fastest method on operations $A \cdot v$ and $v \cdot A$. However, as we will show shortly in \S~\ref{sec:mgd_runtimes}, it has negligible effect in the context of overall ML training time.

\subsection{End-to-End MGD Runtimes} \label{sec:mgd_runtimes}
\vspace{1mm}
In this subsection, we discuss the end-to-end MGD runtime performance with different compression schemes.

\vspace{1mm}
\noindent \textbf{Compared Methods.}
We compare TOC with DEN, CSR, CVI, DVI, Snappy, and Gzip in C++ implementation.
We also integrate TOC into Bismarck and compare it with DEN and CSR implemented in Bismarck, ScikitLearn, and TensorFlow. They are
denoted as ML system name plus data format, e.g., BismarckTOC, ScikitLearnDEN, and TensorFlowCSR.\\
\noindent \textbf{Machine Learning Models.} 
We choose three ML models: Logistic regression (LR), Support vector machine (SVM), and Neural network (NN).
LR/SVM/NN use the standard logistic/hinge/cross-entropy loss respectively. For LR and SVM, we use the standard one-versus-the-other technique
to do the multi-class classification. Our NN has a feed-forward structure with two hidden layers
of 200 and 50 neurons using the sigmoid activation function, and the output layer activation function is sigmoid
for binary output and softmax for multi-class outputs. For Mnist, the output has 10 classes, while all the other datasets have
binary outputs. \\
\noindent \textbf{MGD Training.} We use MGD to train the ML models. Each dataset is divided into mini-batches with 250 rows encoded with different methods.
For the sake of simplicity, we run MGD for fixed 10 epochs. The results of more sophisticated termination conditions are similar.
In each epoch, we visit all the mini-batches and update ML models using each mini-batch. For SVM/LR, we train sequentially. For NN, we use the classical way~\cite{dean2012large} to train the network parallelly. The end-to-end MGD runtimes include all the epochs of training time but do NOT include the compression time because in practice it is a one-time cost and is typically amortized among different ML models.

\vspace{1mm}
\noindent \textbf{Dataset Generation.}  We use the same technique reported in \cite{elgohary2016compressed} to generate scaled real datasets, e.g., Imagenet1m (1 million rows) and Mnist25m (25 million rows).

\vspace{1mm}
\noindent \textbf{Summary of Results.} Table~\ref{tab:model_runtimes} presents the overall results on datasets Imagenet and Mnist. We put the results on the remaining datasets to Appendix~\ref{appendix:mgd_runtimes} for brevity.

On Imagenet1m and Mnist1m, mini-batches encoded using all the methods fit into memory. In this case, CVI and TOC
are the fastest methods. General compression schemes like Snappy and Gzip are much slower due to their significant decompression overheads.
It is interesting to see that TOC is even faster than CVI for LR and SVM on ImageNet1m despite
the fact that matrix operations of TOC are slower on ImageNet1m and all the data fit into memory. The reason is that TOC
reduces the initial IO time because of its better compression ratio. For example, TOC uses 10 seconds to read the data
while CVI takes 36 seconds to read the data on ImageNet1m. On Mnist1m, CVI is faster than TOC for LR and SVM because
we need to train ten LR/SVM models and there are more matrix operations involved.

On Imagenet25m and Mnist25m, only mini-batches encoded using Snappy, Gzip, and TOC fit into memory.
In this case, TOC is up to 1.4x/5.6x/4.8x faster than the state-of-the-art methods for NN/LR/SVM respectively.
The speed-up of TOC for LR/SVM is larger on Imagenet25m than Mnist25m, as Mnist25m has ten output classes and we train ten models so there are more matrix operations involved.

\begin{table}[th!]
 \vspace{1mm}
 \caption{End-to-end MGD runtimes (in minutes) for training machine learning models: Neural network (NN),
 Logistic regression (LR), and Support vector machine (SVM) on datasets Imagenet and Mnist.
  Imagenet1m and Imagenet25m are 7GB and 170GB respectively, while Mnist1m and Mnist25m are 6GB and 150GB respectively.}
 \label{tab:model_runtimes}
 \centering
       \begin{tabular}{|c||c|c|c|c|c|c|c|c|c|c|c|c|}
         \hline
         \multirow{2}{*}{\textbf{Methods}} & \multicolumn{3}{c|}{\textbf{Imagenet1m}} &
         \multicolumn{3}{c|}{\textbf{Imagenet25m}} & \multicolumn{3}{c|}{\textbf{Mnist1m}}
         & \multicolumn{3}{c|}{\textbf{Mnist25m}} \\
            & NN & LR & SVM & NN & LR & SVM  & NN & LR & SVM  & NN & LR & SVM \\
         \hline
         \hline
         TOC (ours) & 12.3 & \textbf{0.7} & \textbf{0.7} & \textbf{249} & \textbf{13} & \textbf{13} & \textbf{9.0} & 2.1 & 2.1 & \textbf{182} & \textbf{52} & \textbf{54} \\
         \hline
         DEN & 14.6 & 3.9 & 3.8 & 666 &  374 & 360 & 15.8 & 7.9 & 7.8 & 708 & 526 & 545 \\
         \hline 
         CSR & 12.7 & 2.1 &  2.1 & 428 & 199 & 187 & 10.8 & 1.6 & 1.6 & 346 & 156 & 155 \\
         \hline
         CVI & 12.5 & 1.0 & 1.1 & 323 &  98 & 83 & 9.6 & \textbf{1.4} & \textbf{1.4} & 250 & 92 & 91.6 \\
         \hline
         DVI & 13.0 & 1.2 & 1.2 & 311 & 73.1 & 63 & 14.5 & 6.2 & 6.4 & 385 & 224 & 226\\
         \hline
         Snappy & 14.8 & 3.9 & 4.0 & 348 & 126 & 127 & 15.8 & 8.5 & 8.4 & 363 & 210 & 213 \\
         \hline
         Gzip & 20.8 & 11.7 & 12.5 & 463 & 247 & 255 & 20.5 & 12.6 & 12.9 & 393 & 238 & 243 \\
         \hline
         \hline
         BismarckTOC & 12.6 & 0.76 & 0.77 & 264 & 13.8 & 13.7 & 10.3 & 2.2 & 2.2 & 198 & 54 & 57 \\
         \hline
         BismarckDEN & N/A & 3.5 & 3.2 & N/A & 309 & 310 & N/A & 7.2 & 7.1 & N/A & 428 & 421 \\
         \hline
         BismarckCSR & N/A & 2.4 & 2.2 & N/A & 141 & 134 & N/A & 1.8 & 1.7 & N/A & 114 & 110 \\
         \hline
         ScikitLearnDEN & 14.7 & 4 & 3.6 & 633 & 454 & 456 & 14.8 & 8.1 & 7.2 & 638 & 536 & 488 \\
         \hline
         ScikitLearnCSR & 42.7 & 2.4 & 2.2 & 1003 & 332 & 334 & 32.9 & 4.4 & 3.3 & 865 & 303 & 284 \\
          \hline
         TensorFlowDEN & \textbf{11.2} & 3.6 & 3.4 & 550 & 426  & 439 & 10.9 & 4.4 & 4.2 & 537 & 439 & 427\\
         \hline
         TensorFlowCSR & 18.4 & 4.4 & 4.3 & 601 & 373 & 359 & 14.8 & 6.7 & 6.5 & 512 & 372 & 341 \\
         \hline
     \end{tabular}
\end{table}

\begin{figure}[th!]
\centering
    \includegraphics[width=0.97\linewidth]{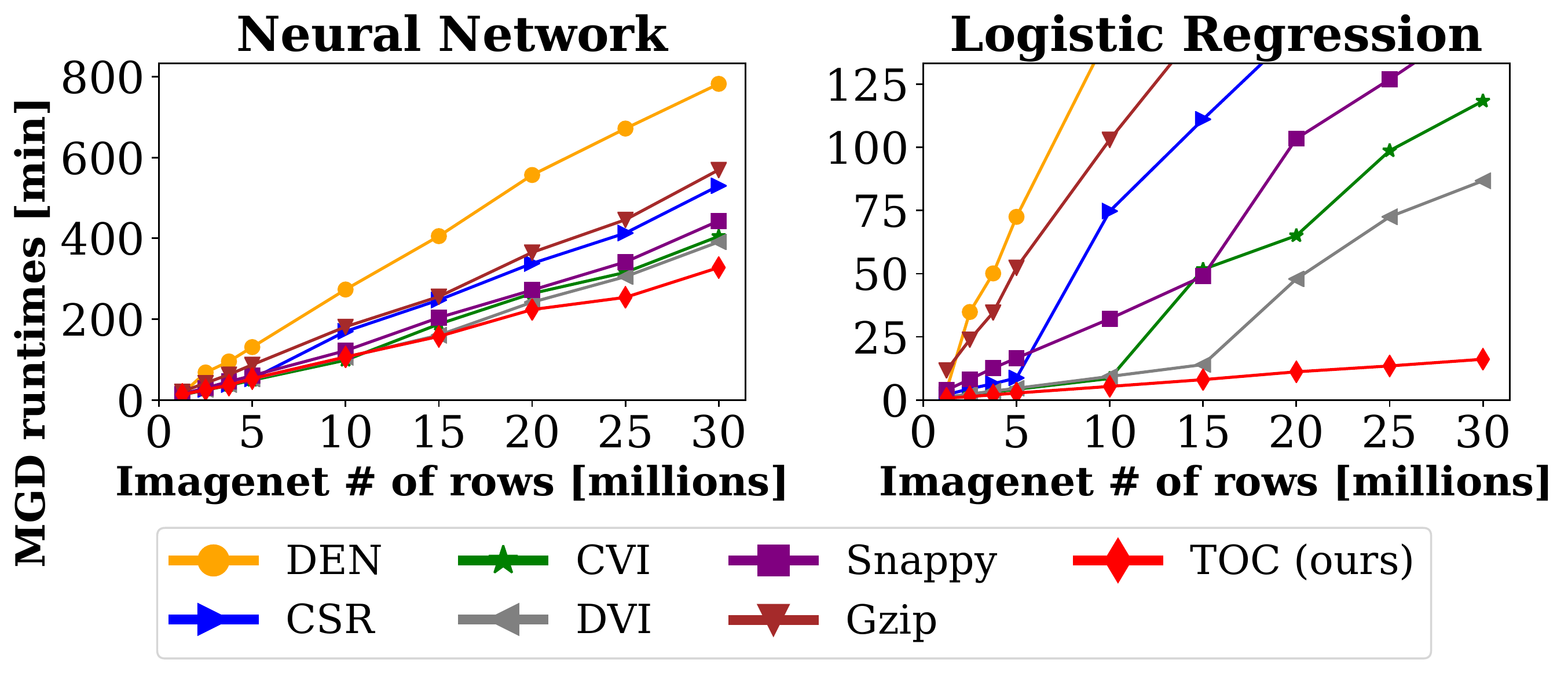}
\vspace{-3mm}
    \caption{End-to-end MGD runtimes of ML training.}
\label{fig:model_runtimes}
\vspace{-3mm}
\end{figure}

\noindent \textbf{More Dataset Sizes.} We also study the MGD runtime over more different
dataset sizes. Figure~\ref{fig:model_runtimes} presents the results. In general, TOC remains the fastest method among all the settings we have tested.
When the dataset is small, CSR, CVI, and DVI have comparable performance to TOC because all the data fit into memory.
When the dataset is large, TOC is faster than other methods because only TOC, Gzip, and Snappy data fit into memory and TOC avoids the decompression.
The speed-up of TOC is larger on Logistic regression than on Neural network because there are more matrix operations involved in training Neural network.

\begin{figure}[th!]
\centering
    \includegraphics[width=0.97\linewidth]{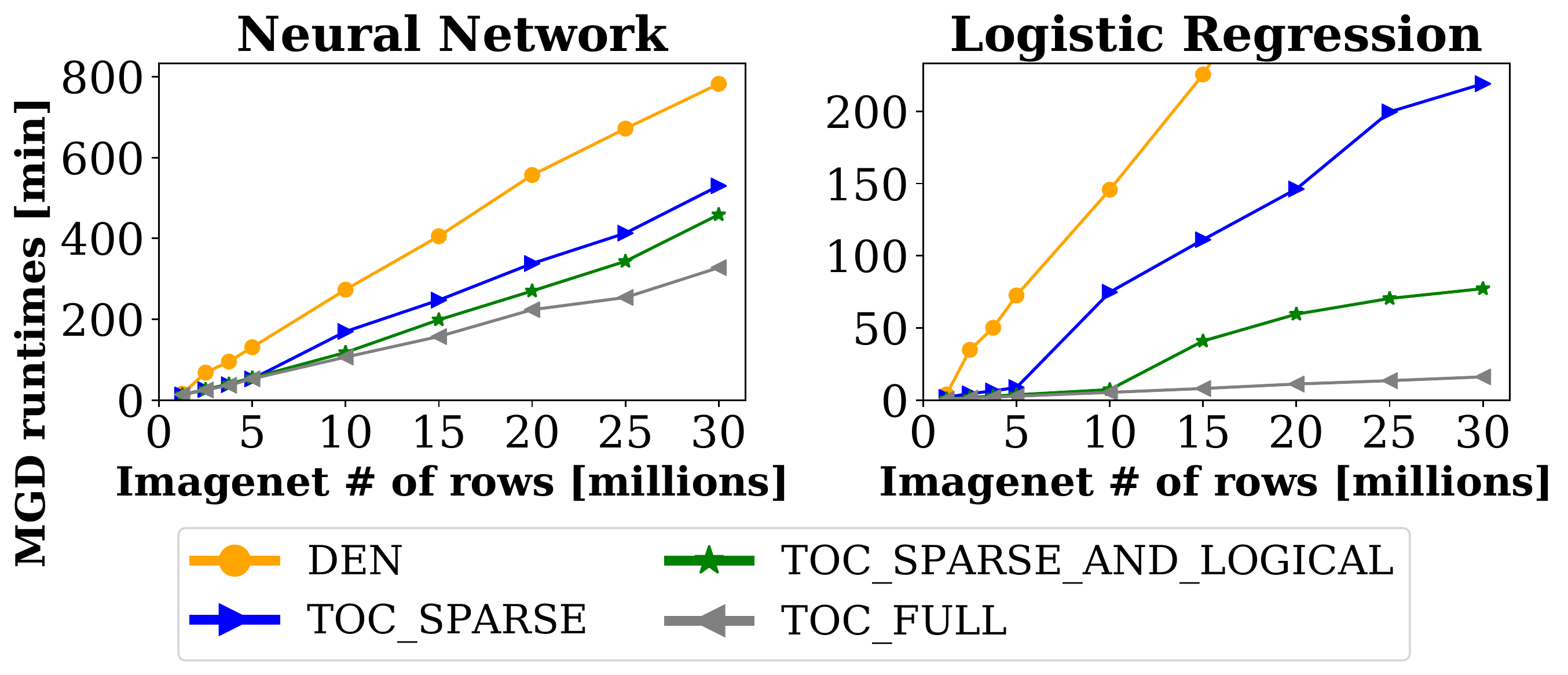}
\vspace{-3mm}
    \caption{Ablation study of TOC for MGD runtimes.}
\label{fig:model_runtimes_ablation_study}
\vspace{-3mm}
\end{figure}

\noindent \textbf{Ablation Study.} We conduct an ablation study to verify whether the components in Figure~\ref{fig:encoding-overview} actually matter for TOC's performance in reducing MGD runtimes. Specifically, we compare three variants of TOC: TOC\_SPARSE (sparse encoding), TOC\_SPARSE\_AND\_LOGICAL (sparse and logical encoding), and TOC\_FULL (all the encoding techniques). Figure~\ref{fig:model_runtimes_ablation_study} presents the results. With more 
encoding techniques used, TOC's performance becomes better, which shows the effectiveness of all our encoding components.

\begin{figure*}[!htbp]
\centering
    \includegraphics[width=0.97\linewidth]{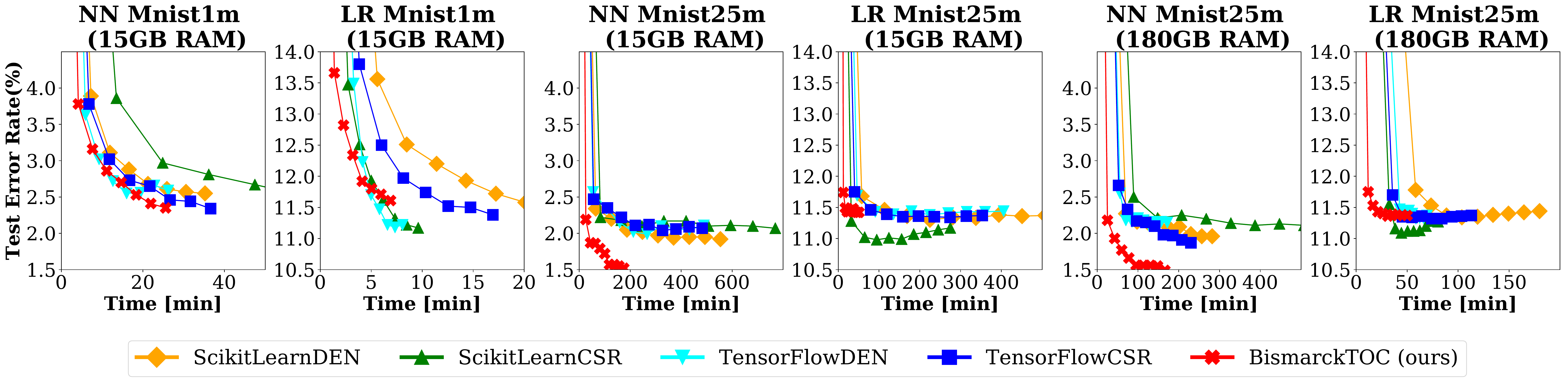}
\vspace{-2mm}
    \caption{Test error rate on Mnist dataset as a function of time on different systems.}
\label{fig:error_time}
\vspace{-3mm}
\end{figure*}

\noindent \textbf{Comparisons with Popular Machine Learning Systems.}
Table~\ref{tab:model_runtimes} also includes the MGD runtimes of DEN and CSR in Bismarck, ScikitLearn, and TensorFlow.
We change the code of using TensorFlow and ScikitLearn a bit so that they can do disk-based learning when the dataset does
not fit into memory. The table also includes BismarckTOC, which typically
has less than 10 percent overhead compared with running TOC in our c++ program. This overhead is caused by the
fudge factor of the database storage thus a bit larger disk IO time. On Imagenet1m and Mnist1m, BismarckTOC is comparable with the best methods in these systems (TensorFlowDEN) for NN but up to 3.2x/2.9x faster than the best methods in these systems for LR/SVM
respectively. On Imagenet25m and Mnist25m, BismarckTOC is up to
2.6x/10.2x/9.8x faster than the best methods in other systems for NN/LR/SVM respectively because only the TOC data fit into memory. Thus, integrating TOC into these ML systems can greatly benefit their MGD performance.\\
\vspace{-0.1mm}
\noindent \textbf{Accuracy Study.}
We also plot the error rate of neural network and logistic regression as a function of time on Mnist.
The goal is to compare the convergence rate of BismarckTOC with other standard tools like ScikitLearn and TensorFlow.
For Mnist1m (7GB) and Mnist25m (170GB), we train 30 epochs and 10 epochs, respectively.
Figure~\ref{fig:error_time} presents the results. On Mnist1m and a 15GB RAM machine, BismarckTOC and TensorFlowDEN finished the training roughly at the same time, this verified our
claim that BismarckTOC has comparable performance with the state-of-the-art ML system if the data fit into memory.
On Mnist25m and a 15GB RAM machine, BismarckTOC finished the training much faster than other ML systems because only TOC data fit into memory.
We also used a machine with 180GB RAM on Mnist25m which did not change BismarckTOC's performance but boosted the performance of TensorFlow and
ScikitLearn to be comparable with BismarckTOC as all the data fit into memory. However, renting a 180GB RAM machine is more expensive than renting a 15GB RAM machine.
Thus, we believe BismarckTOC can significantly reduce users' cloud cost.

\subsection{Compression and Decompression Runtimes} \label{compression_and_decompression_runtimes}
\begin{figure}[th!]
\centering
    \includegraphics[width=0.97\linewidth]{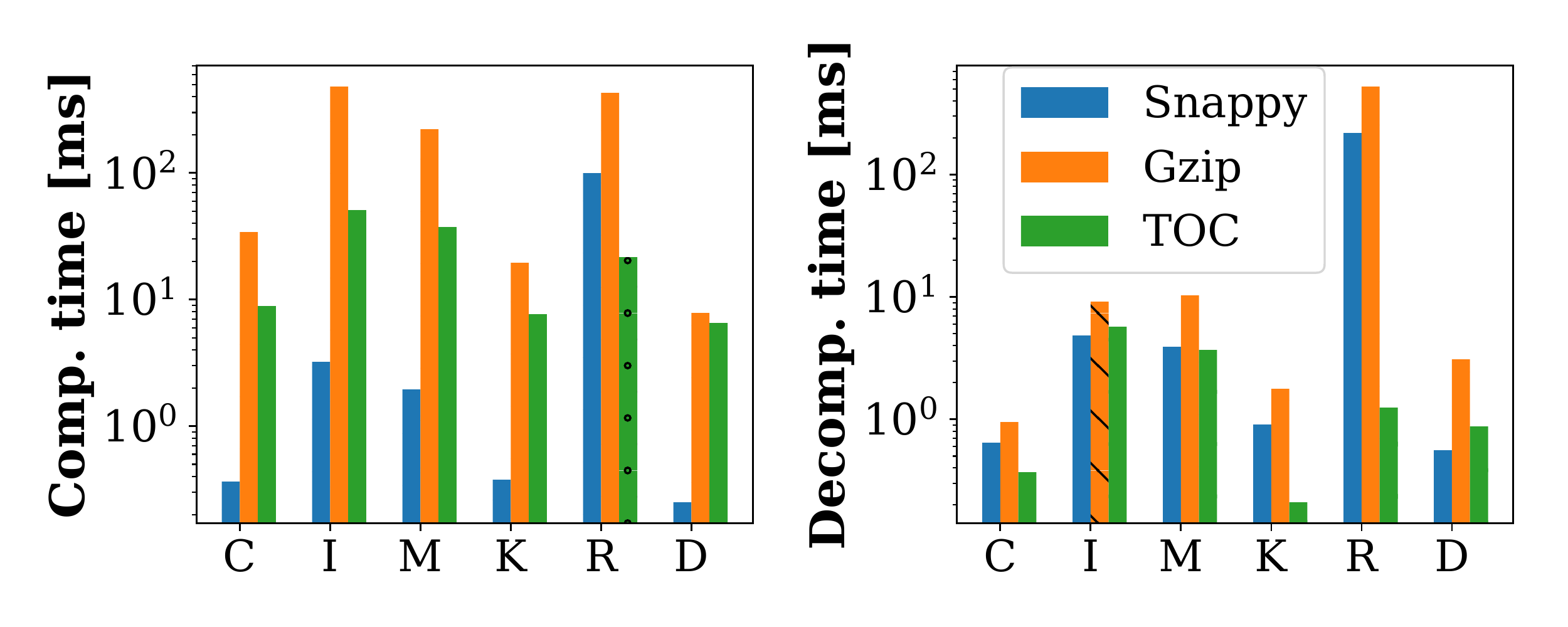}
\vspace{2mm}
    \caption{\textbf{Left:} Compression time of Snappy, Gzip and TOC on a mini-batch with 250 rows. \textbf{Right:} Decompression time
    of Snappy, Gzip, and TOC on a mini-batch with 250 rows.}
\label{fig:compression_and_decompression_runtime}
\vspace{-3mm}
\end{figure}

\vspace{1mm}
We measured the compression and decompression time of Snappy, Gzip, and TOC on mini-batches with 250 rows. The results are similar for 
other mini-batch sizes. Figure~\ref{fig:compression_and_decompression_runtime} presents the results. TOC is faster than Gzip but slower than Snappy for compression. However, TOC is faster than both Gzip and Snappy
for decompression.

\section{Discussion} \label{sec:discuss}
\noindent \textbf{Advanced Neural Network.}
It is possible to apply TOC to more advanced neural networks such as convolutional neural networks on images. One just need
to apply the common image-to-column~\cite{lai2018cmsis} operation, which replicates the pixels of each sliding window as a matrix column.
This way, the convolution operation can be expressed as the matrix multiplication operation over the replicated matrix.
The replicated matrix can be compressed by TOC and we expect higher compression ratios due to the data replication.

\section{Related Work} \label{sec:related}


\noindent \textbf{Data Compression for Analytics.}
There is a long line of research~\cite{abadi2006integrating,raman2013db2,li2013bitweaving,wesley2014leveraging,elgohary2016compressed,wang2017experimental} of integrating data compression into databases and relational query processing workloads on the compressed data. TOC is orthogonal to these works since TOC focuses on a different workload---mini-batch stochastic gradient descent of machine learning training.

\noindent \textbf{Machine Learning Analytics Systems.}
There are a number of systems (e.g., MLib~\cite{meng2016mllib}, MadLib~\cite{hellerstein2012madlib}, Systemml~\cite{boehm2014hybrid,elgohary2016compressed}, Bismarck~\cite{bismarck}, SimSQL~\cite{cai2013simulation}, ScikitLearn~\cite{scikit-learn}, MLBase~\cite{kraska2013mlbase}, and TensorFlow~\cite{abadi2016tensorflow}) for machine learning workloads. Our work focuses on the algorithm perspective and is complementary to these systems, i.e., integrating TOC into these systems can greatly benefit their ML training performance.

\noindent \textbf{Compressed Linear Algebra (CLA). }
CLA~\cite{elgohary2016compressed} compresses the whole dataset and applies batch gradient descent related operations such as vanilla BGD, L-BFGS, and conjugate gradient methods,
while TOC focuses on MGD. Furthermore, CLA needs to store an explicit dictionary. When applying CLA to BGD,
there are many references to dictionary entries so the dictionary cost is amortized. On a small mini-batch,
there are not that many references to the dictionary entries so the explicit dictionary cost makes the CLA compression
ratio less desirable. On the contrary, TOC is adapted from LZW and it does not store an explicit dictionary, so TOC
achieves good compression ratios even on small mini-batches.

\noindent\textbf{Factorized Learning.}
Factorized machine learning techniques~\cite{orion,olteanuf,kumar2016join,chen2017towards} push machine learning computations through joins
and avoid the schema-based redundancy on denormalized datasets. These techniques need a schema to define the 
static redundancies in the denormalized datasets, while TOC can find the redundancies in the datasets 
automatically without a schema. Furthermore, factorized learning techniques work for BGD while TOC focuses on MGD.

\section{Conclusion and Future Work} \label{sec:conclusion}
Mini-batch stochastic gradient descent (MGD) is a workhorse algorithm of modern ML. In this paper, we propose a lossless data compression scheme called tuple-oriented compression (TOC) to reduce memory/storage footprints and runtimes for MGD. TOC follows a design principle that tailors the compression scheme to the data access pattern of MGD in a way that preserves row/column boundaries in mini-batches and adapts matrix operation executions to the compression scheme as much as possible. This enables TOC to attain both good compression ratios and decompression-free executions
for matrix operations used by MGD.
There are a number of interesting directions for future work, including determining more workloads that can execute directly on TOC outputs and investigating the common structures between the adaptable workloads and compression schemes.

\section*{Acknowledgments}
We thank all the anonymous reviewers. This work was partially supported by a gift from Google.

{\small
  \bibliographystyle{abbrv}
  \bibliography{paper}

\begin{thebibliography}{10}

\bibitem{abadi2006integrating}
D.~Abadi, S.~Madden, and M.~Ferreira.
\newblock Integrating compression and execution in column-oriented database
  systems.
\newblock In {\em Proceedings of the 2006 ACM SIGMOD international conference
  on Management of data}, pages 671--682. ACM, 2006.

\bibitem{abadi2016tensorflow}
M.~Abadi et~al.
\newblock Tensorflow: A system for large-scale machine learning.
\newblock In {\em OSDI}, volume~16, pages 265--283, 2016.

\bibitem{amini2009learning}
M.~Amini, N.~Usunier, and C.~Goutte.
\newblock Learning from multiple partially observed views-an application to
  multilingual text categorization.
\newblock In {\em Advances in neural information processing systems}, pages
  28--36, 2009.

\bibitem{ashari2015optimizing}
A.~Ashari et~al.
\newblock On optimizing machine learning workloads via kernel fusion.
\newblock In {\em ACM SIGPLAN Notices}, volume~50, pages 173--182. ACM, 2015.

\bibitem{babenko2016efficient}
A.~Babenko and V.~Lempitsky.
\newblock Efficient indexing of billion-scale datasets of deep descriptors.
\newblock In {\em Proceedings of the IEEE Conference on Computer Vision and
  Pattern Recognition}, pages 2055--2063, 2016.

\bibitem{bengio2012practical}
Y.~Bengio.
\newblock Practical recommendations for gradient-based training of deep
  architectures.
\newblock In {\em Neural networks: Tricks of the trade}, pages 437--478.
  Springer, 2012.

\bibitem{blelloch2001introduction}
G.~E. Blelloch.
\newblock Introduction to data compression.
\newblock {\em Computer Science Department, Carnegie Mellon University}, 2001.

\bibitem{boehm2014hybrid}
M.~Boehm et~al.
\newblock Hybrid parallelization strategies for large-scale machine learning in
  systemml.
\newblock {\em Proceedings of the VLDB Endowment}, 7(7):553--564, 2014.

\bibitem{cai2013simulation}
Z.~Cai et~al.
\newblock Simulation of database-valued markov chains using simsql.
\newblock In {\em Proceedings of the 2013 ACM SIGMOD International Conference
  on Management of Data}, pages 637--648. ACM, 2013.

\bibitem{chelba2013one}
C.~Chelba et~al.
\newblock One billion word benchmark for measuring progress in statistical
  language modeling.
\newblock {\em arXiv preprint arXiv:1312.3005}, 2013.

\bibitem{chen2017towards}
L.~Chen et~al.
\newblock Towards linear algebra over normalized data.
\newblock {\em Proceedings of the VLDB Endowment}, 10(11):1214--1225, 2017.

\bibitem{dean2009challenges}
J.~Dean.
\newblock Challenges in building large-scale information retrieval systems:
  invited talk.
\newblock In {\em Proceedings of the Second ACM International Conference on Web
  Search and Data Mining}, pages 1--1. ACM, 2009.

\bibitem{dean2012large}
J.~Dean et~al.
\newblock Large scale distributed deep networks.
\newblock In {\em Advances in neural information processing systems}, pages
  1223--1231, 2012.

\bibitem{elgohary2016compressed}
A.~Elgohary et~al.
\newblock Compressed linear algebra for large-scale machine learning.
\newblock {\em Proceedings of the VLDB Endowment}, 9(12):960--971, 2016.

\bibitem{bismarck}
X.~Feng et~al.
\newblock {Towards a Unified Architecture for in-RDBMS Analytics}.
\newblock In {\em Proceedings of the 2012 ACM SIGMOD International Conference
  on Management of Data}, pages 325--336. ACM, 2012.

\bibitem{garavaglia1998smart}
S.~Garavaglia and A.~Sharma.
\newblock A smart guide to dummy variables: Four applications and a macro.
\newblock In {\em Proceedings of the Northeast SAS Users Group Conference},
  page~43, 1998.

\bibitem{harnik2012estimation}
D.~Harnik et~al.
\newblock Estimation of deduplication ratios in large data sets.
\newblock In {\em Mass Storage Systems and Technologies (MSST), 2012 IEEE 28th
  Symposium on}, pages 1--11. IEEE, 2012.

\bibitem{hellerstein2012madlib}
J.~M. Hellerstein et~al.
\newblock The madlib analytics library: or mad skills, the sql.
\newblock {\em Proceedings of the VLDB Endowment}, 5(12):1700--1711, 2012.

\bibitem{hinton2012neural}
G.~Hinton, N.~Srivastava, and K.~Swersky.
\newblock Neural networks for machine learning lecture 6a overview of
  mini-batch gradient descent.
\newblock {\em Cited on}, page~14, 2012.

\bibitem{kaoudi2017cost}
Z.~Kaoudi et~al.
\newblock A cost-based optimizer for gradient descent optimization.
\newblock In {\em Proceedings of the 2017 ACM International Conference on
  Management of Data}, pages 977--992. ACM, 2017.

\bibitem{kourtis2008optimizing}
K.~Kourtis, G.~Goumas, and N.~Koziris.
\newblock Optimizing sparse matrix-vector multiplication using index and value
  compression.
\newblock In {\em Proceedings of the 5th conference on Computing frontiers},
  pages 87--96. ACM, 2008.

\bibitem{kraska2013mlbase}
T.~Kraska et~al.
\newblock Mlbase: A distributed machine-learning system.
\newblock In {\em Cidr}, volume~1, pages 2--1, 2013.

\bibitem{kumar2016join}
A.~Kumar et~al.
\newblock To join or not to join? thinking twice about joins before feature
  selection.
\newblock In {\em Proceedings of the 2016 International Conference on
  Management of Data}, pages 19--34. ACM, 2016.

\bibitem{orion}
A.~Kumar, J.~Naughton, and J.~M. Patel.
\newblock Learning generalized linear models over normalized data.
\newblock In {\em Proceedings of the 2015 ACM SIGMOD International Conference
  on Management of Data}, pages 1969--1984. ACM, 2015.

\bibitem{lai2018cmsis}
L.~Lai, N.~Suda, and V.~Chandra.
\newblock Cmsis-nn: Efficient neural network kernels for arm cortex-m cpus.
\newblock {\em arXiv preprint arXiv:1801.06601}, 2018.

\bibitem{lemire2015decoding}
D.~Lemire and L.~Boytsov.
\newblock Decoding billions of integers per second through vectorization.
\newblock {\em Software: Practice and Experience}, 45(1):1--29, 2015.

\bibitem{li2013bitweaving}
Y.~Li and J.~M. Patel.
\newblock Bitweaving: fast scans for main memory data processing.
\newblock In {\em Proceedings of the 2013 ACM SIGMOD International Conference
  on Management of Data}, pages 289--300. ACM, 2013.

\bibitem{Lichman:2013}
M.~Lichman.
\newblock {UCI} machine learning repository, 2013.

\bibitem{meng2016mllib}
X.~Meng et~al.
\newblock Mllib: Machine learning in apache spark.
\newblock {\em The Journal of Machine Learning Research}, 17(1):1235--1241,
  2016.

\bibitem{mishkin2016systematic}
D.~Mishkin, N.~Sergievskiy, and J.~Matas.
\newblock Systematic evaluation of cnn advances on the imagenet.
\newblock {\em arXiv preprint arXiv:1606.02228}, 2016.

\bibitem{hogwild}
F.~Niu et~al.
\newblock Hogwild: A lock-free approach to parallelizing stochastic gradient
  descent.
\newblock In {\em NIPS}, 2011.

\bibitem{scikit-learn}
F.~Pedregosa et~al.
\newblock Scikit-learn: Machine learning in {P}ython.
\newblock {\em Journal of Machine Learning Research}, 12:2825--2830, 2011.

\bibitem{qin2017scalable}
C.~Qin, M.~Torres, and F.~Rusu.
\newblock Scalable asynchronous gradient descent optimization for out-of-core
  models.
\newblock {\em Proceedings of the VLDB Endowment}, 10(10):986--997, 2017.

\bibitem{raman2013db2}
V.~Raman et~al.
\newblock Db2 with blu acceleration: So much more than just a column store.
\newblock {\em Proceedings of the VLDB Endowment}, 6(11):1080--1091, 2013.

\bibitem{ruder2016overview}
S.~Ruder.
\newblock An overview of gradient descent optimization algorithms.
\newblock {\em arXiv preprint arXiv:1609.04747}, 2016.

\bibitem{russakovsky2015imagenet}
O.~Russakovsky et~al.
\newblock Imagenet large scale visual recognition challenge.
\newblock {\em International Journal of Computer Vision}, 115(3):211--252,
  2015.

\bibitem{saad2003iterative}
Y.~Saad.
\newblock {\em Iterative methods for sparse linear systems}, volume~82.
\newblock siam, 2003.

\bibitem{olteanuf}
M.~Schleich, D.~Olteanu, and R.~Ciucanu.
\newblock Learning linear regression models over factorized joins.
\newblock In {\em Proceedings of the 2016 International Conference on
  Management of Data}, pages 3--18. ACM, 2016.

\bibitem{shai}
S.~Shalev-Shwartz and S.~Ben-David.
\newblock {\em {Understanding Machine Learning: From Theory to Algorithms}}.
\newblock Cambridge University Press, 2014.

\bibitem{SSSSS10}
S.~Shalev-Shwartz et~al.
\newblock Learnability, stability and uniform convergence.
\newblock {\em The Journal of Machine Learning Research}, 11:2635--2670, 2010.

\bibitem{wang2017experimental}
J.~Wang et~al.
\newblock An experimental study of bitmap compression vs. inverted list
  compression.
\newblock In {\em Proceedings of the 2017 ACM International Conference on
  Management of Data}, pages 993--1008. ACM, 2017.

\bibitem{welch1984technique}
T.~A. Welch.
\newblock A technique for high-performance data compression.
\newblock {\em Computer}, 6(17):8--19, 1984.

\bibitem{wesley2014leveraging}
R.~M.~G. Wesley and P.~Terlecki.
\newblock Leveraging compression in the tableau data engine.
\newblock In {\em Proceedings of the 2014 ACM SIGMOD international conference
  on Management of data}, pages 563--573. ACM, 2014.

\bibitem{wiggins2001image}
R.~H. Wiggins et~al.
\newblock Image file formats: Past, present, and future 1.
\newblock {\em Radiographics}, 21(3):789--798, 2001.

\bibitem{wu2017bolt}
X.~Wu et~al.
\newblock Bolt-on differential privacy for scalable stochastic gradient
  descent-based analytics.
\newblock In {\em Proceedings of the 2017 ACM International Conference on
  Management of Data}, pages 1307--1322. ACM, 2017.

\bibitem{yu2012large}
H.-F. Yu et~al.
\newblock Large linear classification when data cannot fit in memory.
\newblock {\em ACM Transactions on Knowledge Discovery from Data (TKDD)},
  5(4):23, 2012.

\bibitem{ziv1977universal}
J.~Ziv and A.~Lempel.
\newblock A universal algorithm for sequential data compression.
\newblock {\em IEEE Transactions on information theory}, 23(3):337--343, 1977.

\bibitem{ziv1978compression}
J.~Ziv and A.~Lempel.
\newblock Compression of individual sequences via variable-rate coding.
\newblock {\em IEEE transactions on Information Theory}, 24(5):530--536, 1978.

\end{thebibliography}
}

\appendix

\section{Proof of Theorems} \label{thoerem_proof}
\vspace{1mm}
\subsection{Theorem~\ref{theorem:matrix_times_vector}} \label{appendix:matrix_times_vector}
\begin{proof}
Without loss of generality, we use a specific row $A[i, :]$ in the proof. First, we substitute $A[i, :]$ with sequences stored in the prefix tree $\mathbf{C'}$, then
\begin{align} \label{equation:matrix_times_vector_substitution}
	A[i, :] \cdot v =\sum_{j=0}^{\mathrm{len}(\mathbf{D}[i, :])-1} \mathbf{C'}[\mathbf{D}[i,j]].seq \cdot v
\end{align}
Plug Equation~\ref{equation:matrix_times_vector_definition} into Equation~\ref{equation:matrix_times_vector_substitution},
we get Equation~\ref{equation:matrix_times_vector_first_equation}.
Following the definition of the sequence of the tree node,
we immediately get Equation~\ref{equation:vector_times_matrix_second_equation}.
\end{proof}

\subsection{Theorem~\ref{theorem:vector_times_matrix}}
\label{appendix:vector_times_matrix}
\begin{proof}
We substitute $A$ with sequences stored in $\mathbf{C'}$
\begin{align}
	v \cdot A & = \sum_{i=0}^{n-1} v[i] \cdot A[i, :] \nonumber \\
          & = \sum_{i=0}^{n-1} \sum_{j=0}^{\mathrm{len}(\mathbf{D}[i, :])-1} v[i] \cdot \mathbf{C'}[\mathbf{D}[i][j]].seq.
          \label{equation:vector_times_matrix_substitution}
\end{align}
Merge terms in Equation~\ref{equation:vector_times_matrix_substitution} with same sequences
\begin{align}
	v \cdot A = \sum_{i=1}^{\mathrm{len}(\mathbf{C'})-1} \mathbf{C'}[i].seq \cdot (\sum_{\substack{\mathbf{D}[i_k, j_k]=i,
    					\forall i_k \in \aleph,
                        \forall j_k \in \aleph}}
                        v[i_k]) \label{equation:vector_times_matrix_merge_terms}
\end{align}
Plug Equation~\ref{equation:vector_times_matrix_definition} into Equation~\ref{equation:vector_times_matrix_merge_terms}, we get Equation~\ref{equation:vector_times_matrix_first_equation}.
\end{proof}

\section{More Algorithms} \label{more_algorithms}
\vspace{1mm}
 \subsection{Right Multiplication} \label{appendix:right_multiplication}
 We present how to compute $A\cdot M,$ where $M$ is an uncompressed matrix and $A$ is a 
 compressed matrix. This is an extension of right multiplication with vector in \S~\ref{right_multiplication_vector}. 
 \begin{theorem}
 Let $A \in \Re^{n \times m}$, $M \in \Re^{m \times p}$, \textbf{D} be the output of TOC on $A$,
 $\mathbf{C'}$ be the prefix tree built for decoding, $\mathbf{C'}[i].seq$ be the sequence of the tree node defined in 
 \S~\ref{sec:api}, $\mathbf{C'}[i].key$ be the key of the tree node defined in \S~\ref{sec:build_prefix_tree},
 and $\mathbf{C'}[i].parent$ be the parent index of the tree node defined in \S~\ref{sec:build_prefix_tree}.
 Note that $\mathbf{C'}[i].key$ and $\mathbf{C'}[i].seq$ are both sparse representations of vectors
 (i.e., $\mathbf{C'}[i].key \in \Re^{1 \times m}$ and $\mathbf{C'}[i].seq \in \Re^{1 \times m}$).
 Define function $\mathcal{F}(x): \aleph \rightarrow \Re^{1\times p}$ to be
 
 \begin{align} \label{equation:matrix_times_uncompress_definition}
 	\mathcal{F}(x) = \mathbf{C'}[x].seq \cdot M, x = 1, 2, ..., \mathrm{len}(\mathbf{C'})-1.
 \end{align}
 Then, we have
 \begin{align} \label{equation:matrix_times_uncompress_first_equation}
 A[i, :] \cdot M = & \sum_{j=0}^{\mathrm{len}(\mathbf{D}[i, :])-1} \mathcal{F}(\mathbf{D}[i][j]), i = 0, 1, ..., n-1.
 \end{align}
 \end{theorem}
 \begin{proof}
 Without loss of generality, we use a specific row $A[i, :]$ in the proof. First, we substitute $A[i, :]$ with sequences stored in prefix tree $\mathbf{C'}$, then
 \begin{align} \label{equation:matrix_times_uncompress_substitution}
 	A[i, :] \cdot M =\sum_{j=0}^{\mathrm{len}(\mathbf{D}[i, :])-1} \mathbf{C'}[\mathbf{D}[i,j]].seq \cdot M
 \end{align}
 Plug Equation~\ref{equation:matrix_times_uncompress_definition} into Equation~\ref{equation:matrix_times_uncompress_substitution},
 we get Equation~\ref{equation:matrix_times_uncompress_first_equation}.
 \end{proof}
 
 Algorithm~\ref{alg:matrix_times_uncompress_matrix} shows the details.
 First, we scan $\mathbf{C'}$ similar to right multiplication with vector, and we use \textbf{H}[$i$,:] to remember the computed
 value of $\mathcal{F}(i)$.
 
 Second, we scan \textbf{D} to compute $A \cdot M$ stored in \textbf{R}. For $k$th column of the result \textbf{R} and each \textbf{D}[$i$][$j$], we simply add 
 \textbf{H}[\textbf{D}[$i$][$j$]][$k$] to \textbf{R}[$i$][$k$]. Because \textbf{H}[\textbf{D}[$i$][$j$]][$k$] is a random access of \textbf{H}, we let the loop of going over each column be the most inner
 loop so that we can scan \textbf{D} only once and have better cache performance.

\begin{algorithm} [th!]
\caption{Execute $A \cdot M$ on the TOC output.} \label{alg:matrix_times_uncompress_matrix}
\begin{algorithmic}[1]
\Function{MatrixTimesUncompressedMatrix}{\textbf{D}, \textbf{I}, $M$}
	\State \textbf{inputs:} column\_index:value pairs in the first layer of
    \textbf{I}, encoded table \textbf{D}, and uncompressed matrix $M$
    \State \textbf{outputs:} the result of $A \cdot M$ in \textbf{R}
    \State $\mathbf{C'} \gets$ \Call{BuildPrefixTree}{\textbf{I}, \textbf{D}}
    \State \textbf{H} $\gets \left[0\right]$ \Comment{initialize as a zero matrix}
    \For {$i$ = 1 to len($\mathbf{C'}$)-1} \Comment{scan $\mathbf{C'}$ to compute \textbf{H}}
        \For {$j$ = 0 to num\_of\_columns($M$)-1}
    	\State \textbf{H}[$i][j]  \gets \mathbf{C'}[i].key \cdot M[:,j] + \textbf{H}[\mathbf{C'}[i].parent][j$]
        \EndFor
    \EndFor
    \State \textbf{R} $\gets \left[0\right]$ \Comment{initialize as a zero matrix}
    \For {$i$ = 0 to len(\textbf{D})-1} \Comment{scan \textbf{D} to compute \textbf{R}}
	    \For {$j$ = 0 to len(\textbf{D}[$i$,:])-1}
    		\For {$k$ = 0 to num\_of\_columns($M$)-1}
        	\State $\textbf{R}[i][k] \gets \textbf{R}[i][k] + \textbf{H}[\textbf{D}[i][j]][k]$
            \EndFor
        \EndFor
    \EndFor
    \State \textbf{return(R)}
\EndFunction
\end{algorithmic}
\end{algorithm}

\subsection{Left Multiplication}\label{appendix:left_multiplication}
 We discuss how to compute $M\cdot A$ where $M$ is an uncompressed matrix and $A$ is
 a compressed matrix. This is an extension of left multiplication with vector in \S~\ref{left_multiplication_vector}. 
 
 \begin{theorem}
 Let $A \in \Re^{n \times m}$, $M \in \Re^{p \times n}$, \textbf{D} be the output of TOC on $A$,
 $\mathbf{C'}$ be the prefix tree built for decoding, $\mathbf{C'}[i]$.seq be the sequence of the tree node defined in 
 \S~\ref{sec:api}, $\mathbf{C'}[i].key$ be the key of the tree node defined in \S~\ref{sec:build_prefix_tree},
 and $\mathbf{C'}[i].parent$ be the parent index of the tree node defined in \S~\ref{sec:build_prefix_tree}.
 Note that $\mathbf{C'}[i].key$ and $\mathbf{C'}[i].seq$ are both sparse representations of vectors
 (i.e., $\mathbf{C'}[i].key \in \Re^{1 \times m}$ and $\mathbf{C'}[i].seq \in \Re^{1 \times m}$).
 Define function $\mathcal{G}(x): \aleph \rightarrow \Re^{p \times 1}$ to be
 
 \begin{align}
 	\mathcal{G}(x) = \sum_{\substack{\mathbf{D}[i, j]=x,
     					\forall i \in \aleph,
                         \forall j \in \aleph}}
                         M[:,i],~x = 1, 2, ..., \mathrm{len}(\mathbf{C'})-1. \label{equation:uncompress_times_matrix_definition}
 \end{align}
 Then, we have
 
 \begin{align}
 	M \cdot A =& \sum_{i=1}^{\mathrm{len}(\mathbf{C'})-1} \mathbf{C'}[i].seq \cdot \mathcal{G}(i).
     \label{equation:uncompress_times_matrix_first_equation}
 \end{align}
 \end{theorem}
 
 \begin{proof}
 We substitute $A$ with sequences stored in $\mathbf{C'}$
 \begin{align}
 	M \cdot A & = \sum_{i=0}^{n-1} M[:,i] \cdot A[i, :] \nonumber \\
           & = \sum_{i=0}^{n-1} \sum_{j=0}^{\mathrm{len}(\mathbf{D}[i, :])-1} M[:,i] \cdot \mathbf{C'}[\mathbf{D}[i][j]].seq.
           \label{equation:uncompress_times_matrix_substitution}
 \end{align}
 Merge terms in Equation~\ref{equation:uncompress_times_matrix_substitution} with same sequences
 \begin{align}
 	M \cdot A = \sum_{i=1}^{\mathrm{len}(\mathbf{C'})-1} \mathbf{C'}[i].seq \cdot (\sum_{\substack{\mathbf{D}[i_k, j_k]=i,
     					\forall i_k \in \aleph,
                         \forall j_k \in \aleph}}
                         M[:,i_k]) \label{equation:uncompress_times_matrix_merge_terms}
 \end{align}
 Plug Equation~\ref{equation:uncompress_times_matrix_definition} into Equation~\ref{equation:uncompress_times_matrix_merge_terms}, we get Equation~\ref{equation:uncompress_times_matrix_first_equation}.
 \end{proof}
 Algorithm~\ref{alg:uncompress_matrix_times_matrix} shows the details. First, we similarly scan \textbf{D} as left multiplication with vector. Specifically, we initialize $\mathbf{H}$ as a zero matrix,
 and then add $M[k][i]$ to $\mathbf{H}[\mathbf{D}[i][j]][k]$ for each $\mathbf{D}[i][j]$. Note that the \textbf{H} here is stored in transposed manner so that we only need to scan $\mathbf{D}$ once and have good cache performance at the same time.
 
 Second, we scan $\mathbf{C'}$ backwards to actually compute $M \cdot A$ stored in \textbf{R}. Specifically,
 for $i$th row and each $\mathbf{C'}[j]$, we add $\mathbf{C'}[j]$.key * $\mathbf{H}[i][j]$ to the result of $i$th row \textbf{R}[i,:] and add $\mathbf{H}[i][j]$ to
 $\mathbf{H}[i][\mathbf{C'}[j]$.parent].

\begin{algorithm} [th!]
\caption{Execute $M \cdot A$ on the TOC output.}
\label{alg:uncompress_matrix_times_matrix}
\begin{algorithmic}[1]
\Function{UncompressedMatrixTimesMatrix}{\textbf{D}, \textbf{I}, $M$}
	\State \textbf{inputs:} column\_index:value pairs in the first layer of
    \textbf{I}, encoded table \textbf{D}, and uncompressed matrix $M$
    \State \textbf{outputs:} the result of $M \cdot A$ in \textbf{R}
    \State $\mathbf{C'} \gets$ \Call{BuildPrefixTree}{\textbf{I}, \textbf{D}}
    \State \textbf{H} $\gets \left[0\right]$ \Comment{initialize as a zero matrix}
    \For {$i$ = 0 to len(\textbf{D})-1} \Comment{scan \textbf{D} to compute \textbf{H}}
    \For {$j$ = 0 to len(\textbf{D}[$i$,:]) -1}
    \For {$k$ = 0 to num\_of\_rows($M$) -1}
        	\State $\textbf{H}[\textbf{D}[i][j]][k] \gets M[k][i] + \textbf{H}[\textbf{D}[i][j]][k]$
            \EndFor
        \EndFor
    \EndFor
    \State \textbf{R} $\gets \left[0\right]$ \Comment{initialize as a zero matrix}
    \For {$i$ = len($\mathbf{C'}$) -1 to 1} \Comment{scan $\mathbf{C'}$ to compute \textbf{R}}
    \For{$j$ = 0 to num\_of\_rows($M$) -1}
    	\State $\textbf{R}[j,:] \gets \textbf{R}[j,:] + \mathbf{C'}[i]$.key * $\textbf{H}[i][j]$
      \State Add $\textbf{H}[i][j]$ to  $\textbf{H}[\mathbf{C'}[i]$.parent$][j]$
    \EndFor
    \EndFor
    \State \textbf{return(R)}
\EndFunction
\end{algorithmic}
\end{algorithm}

\section{More Time Complexity Analysis} \label{more_complexity_analysis}
Executing $M \cdot A$ and $A \cdot M$ needs to build $\mathbf{C'}$, scan $\mathbf{C'}$,
and scan \textbf{D}. As shown in Algorithm~\ref{alg:build_prefix_tree_c_prime}, building $\mathbf{C'}$ has
complexity $\mathcal{O}(|\mathbf{I}| + |\mathbf{D}|)$ and $|\mathbf{C'}| = |\mathbf{I}| + |\mathbf{D}|$.
When scanning $\mathbf{C'}$ and $\mathbf{D}$, each element needs to do r$ow\_size(\mathbf{M})/col\_size(\mathbf{M})$
operations for $M \cdot A$/$A \cdot M$ respectively. Thus, the time complexity for $M \cdot A$ and $A \cdot M$
is $\mathcal{O}(row\_size(\mathbf{M}) * (|\mathbf{I}| + |\mathbf{D}|))$ and
$\mathcal{O}(col\_size(\mathbf{M}) * (|\mathbf{I}| + |\mathbf{D}|))$ respectively.

%

\section{More Experiments} \label{more_experiments}
\vspace{1mm}
\subsection{Integration TOC into Bismarck}~\label{appendix:integration}
We integrated TOC into Bismarck and replaced its existing matrix kernels.
There are three key parts of the integration. First, we allocate an arena space in
Bismarck shared memory for storing the ML models. Second, we replace the existing
Bismarck matrix kernel with the TOC matrix kernel for updating the ML models.
Third, a database table is used to store the TOC compressed mini-batches and
the serialized bytes of each TOC compressed mini-batch are stored as a bytes field
of variable length in the row. After all these, we modified the UDF
of ML training to read the compressed mini-batch from the table
and use the replaced matrix kernel to update the ML model in the arena.

\begin{table*}[th!]
 \vspace{1mm}
\caption{End-to-end MGD runtimes (in minutes) for training machine learning models: Neural network(NN),
Logistic regression (LR), and Support vector machine (SVM) on datasets Census and Kdd99. Census15m and Census290m is
  7GB and 140GB respectively, while Kdd7m and Kdd200m is 7GB and 200GB respectively.}
\vspace{-3mm}
\label{tab:model_runtimes_census_and_kdd}
\centering
      \begin{tabulary}{1.0\textwidth}{|L||L|L|L|L|L|C|L|L|L|L|L|L|}
        \hline
        \multirow{2}{*}{\textbf{Methods}} & \multicolumn{3}{c|}{\textbf{Census15m}} &
        \multicolumn{3}{c|}{\textbf{Census290m}} & \multicolumn{3}{c|}{\textbf{Kdd7m}}
        & \multicolumn{3}{c|}{\textbf{Kdd200m}} \\
           & NN & LR & SVM & NN & LR & SVM  & NN & LR & SVM  & NN & LR & SVM \\
        \hline
        \hline
        TOC (ours) & \textbf{35} & \textbf{0.8} & \textbf{0.7} & \textbf{702} & \textbf{16} & \textbf{14} & 16.1 & \textbf{0.2} & \textbf{0.2} & \textbf{323} & \textbf{6.1} & \textbf{5.9} \\
        \hline
        DEN & 39 & 4.0 & 4.0 & 1108 &  253 & 251                & 29 & 4.6 & 4.4 & 1003 &  608 & 615 \\
        \hline                                                   
        CSR & 38 & 1.8 &  1.8 & 942 & 161 & 167                 & 19.2 & 0.4 &  0.4 & 438 & 56 & 53 \\
        \hline                                                                                                  
        CVI & 37 & 1.1 & 1.0 & 844 &  80 & 67                   & 18.5 & 0.3 & 0.3 & 422 &  31 & 30 \\
        \hline                                                                                                  
        DVI & 38 & 1.2 & 1.1 & 800 & 46 & 43                    & 28.4 & 1.2 & 1.1 & 611 & 71 & 71 \\
        \hline                                                                                                  
        Snappy & 41 & 4.7 & 4.6 & 905 & 121 & 115              & 27.2 & 3.5 & 3.5 & 616 & 127 & 128  \\
        \hline                                                                                                  
        Gzip & 46 & 11.1 & 11.1 & 965 & 244 & 241               & 33.5 & 7.5 & 7.5 & 683 & 235 & 235  \\
        \hline
        \hline
        BismarckTOC & 38 & 0.87 & 0.88 &  742 & 17.4 & 14.8 & 16.8 & 0.3 & 0.31 & 329 & 6.4 & 6.3 \\
        \hline
        BismarckDEN & N/A & 4.2 & 4.3 & N/A & 321 & 310          & N/A & 4.0 & 3.8 & N/A & 645 & 644 \\
        \hline                                                                                                  
        BismarckCSR & N/A & 3.2 & 3.2 & N/A & 222 & 234          & N/A & 0.9 & 0.9 & N/A & 114 & 115  \\
        \hline                                                                                                  
        ScikitLearnDEN & 73.2 & 7.3 & 6.6 & 1715 & 604 & 580    & 42 & 5 & 4.6 & 1797 & 771 & 772  \\
        \hline                                                                                                  
        ScikitLearnCSR & 105.1 & 5.7 & 5.1 & 2543 & 421 & 408.8   & 44 & 1.7 & 1.5 & 1476 & 166 & 160 \\
        \hline                                                                                                  
        TensorFlowDEN & 38.1 & 9.4 & 10.5 & 1073 & 638 & 610     & 21.4 & 5.5 & 5.1 & 1199 & 781 & 779  \\
        \hline                                                   
        TensorFlowCSR & 54.7 & 15.1 & 14.0 & 1244 & 681 & 661    & \textbf{15.2} & 4.1 & 4.4 & 577 & 300 & 274 \\
        \hline
    \end{tabulary}
\end{table*}

\subsection{End-to-End MGD Runtimes}~\label{appendix:mgd_runtimes}
MGD runtimes on Census and Kdd99 are reported in Table~\ref{tab:model_runtimes_census_and_kdd}. Overall, the results are similar to those presented in \S~\ref{sec:mgd_runtimes}. On small datasets like Census15m and Kdd7m, TOC has comparable performance with other methods. On large datasets like Census290m and Kdd200m, TOC is up to 1.8x/17.8x/18.3x faster than the state-of-the-art compression schemes for NN/LR/SVM respectively. We leave the results of datasets Rcv1 and Deep1Billion because of their extreme sparsity/density such that we do not expect better performance from TOC.

\end{document}